\documentclass[journal]{IEEEtran}

\usepackage{cite}
\usepackage{url}
\usepackage{xcolor}
\usepackage{textcomp}

\ifCLASSINFOpdf
  \usepackage[pdftex]{graphicx}
  \graphicspath{{./figs/}{../02_LIDAR_AD/figs/}}
\else
  \usepackage[dvips]{graphicx}
\fi
\usepackage{algorithm}
\usepackage{algorithmic}

\usepackage[caption=false,font=footnotesize,labelfont=rm,textfont=rm]{subfig}
\usepackage[colorlinks=true, linkcolor=blue, citecolor=blue, urlcolor=blue]{hyperref}

\usepackage{array}
\usepackage{booktabs}
\usepackage{multirow}
\usepackage{makecell}
\usepackage{pifont}

\newcommand{\CYes}{\ding{51}}      % check mark
\newcommand{\CNo}{--}              % absent
\usepackage{amsmath,amssymb,amsfonts}
\usepackage{cuted}       % 长公式跨栏 (strip 环境)

\usepackage{enumitem}
\usepackage{stfloats}    % 底部双栏浮动
\usepackage{ragged2e}    % 文本对齐

\usepackage{tikz}

\begin{document}
\bstctlcite{IEEEtranBSTCTL:no_dash}

% ===========================================================================
% 标题
% ===========================================================================
\title{LIDAR-AD: A Decoder-Free Latent-Interaction Dreamer with Action-Residual Chains for Autonomous Driving}

% ===========================================================================
% 作者栏 — 从 EchoDreamer.tex 提取
% ===========================================================================
\author{Yongzhi~Liu,
  Yang~Xiao,
  Zhong~Cao,
  Zeng~Kang,
  Sunan~Zhang,
  Zhaozhi~Dong,
  Guojun~Yu,
  and~Weichao~Zhuang,~\IEEEmembership{Member,~IEEE}%
  \thanks{This work was supported in part by the Excellent Youth Fund Project of the Jiangsu Basic Research Program under Grant BK20250172, in part by the Major Science and Technology Special Project of Jiangsu Province under Grant BG2025018, and in part by the National Natural Science Foundation of China (NSFC) under Grant 52441204. (Corresponding author: Weichao Zhuang.)}%
  % \thanks{Project page: \url{https://lumen2023.github.io/lidar-ad-website/}.}%
  \thanks{Yongzhi Liu, Yang Xiao, Zeng Kang, Sunan Zhang, Zhaozhi Dong, and Weichao Zhuang are with the School of Mechanical Engineering, Southeast University, Nanjing, Jiangsu 211189, China (e-mail: yongzhiliu@seu.edu.cn; wezhuang@seu.edu.cn).}%
  \thanks{Zhong Cao is with the Department of Civil and Environmental Engineering, University of Michigan, Ann Arbor, MI 48109, USA (e-mail: zhcao@umich.edu).}%
  \thanks{Guojun Yu is with Xheart Technology Co., Ltd., Suzhou, Jiangsu 215100, China (e-mail: gavin.yu@xheart.com).}%
}

% ===========================================================================
% 页眉
% ===========================================================================
\markboth{IEEE Transactions on Knowledge and Data Engineering}%
{Liu \MakeLowercase{\textit{et al.}}: LIDAR-AD: Decoder-Free Latent-Interaction Dreamer}

\maketitle

% ===========================================================================
% 摘要
% ===========================================================================
% 摘要逻辑说明：
% 1. 背景与问题：自动驾驶需要动态交通中的长时序闭环决策，多源观测存在控制无关冗余，而可靠决策依赖风险相关关系和未来动力学。
% 2. 方法：LIDAR-AD 以无解码器潜在对齐、残差动作更新和残差动作序列对比学习改进 latent representation、continuous control 和 long-horizon dynamics。
% 3. 理论与结果：补充 latent-tanh 残差参数化的确定性分析，并总结仿真场景和 nuPlan benchmark 上的实验有效性。
%
% --- 中文翻译 ---
% 自动驾驶需要在动态交通环境中进行长时序闭环决策。
% 潜在世界模型通过在紧凑潜在空间中支持基于想象的决策，为这一问题提供了有效框架。
% 然而，多源驾驶观测包含与控制无关的冗余，而可靠驾驶决策依赖风险相关关系、未来动力学和连续动作调整。
% 这一不匹配使得观测重建和绝对动作建模在学习决策相关潜在动力学时并非最优。
% 本文提出 LIDAR-AD，一种面向自动驾驶、带有动作残差链的无解码器潜在交互 Dreamer。
% LIDAR-AD 以冗余消减潜在对齐替代观测重建，促使模型学习多源驾驶输入中风险相关关系的紧凑表征。
% 它进一步将车辆控制建模为残差动作更新，并利用残差动作序列对比学习，将由多步残差驱动的 rollout 与未来潜在状态对齐。
% 确定性分析表明，latent-tanh 残差参数化在保持内部动作可达性的同时，将平滑长时序控制表示为紧凑局部更新。
% 这些设计共同提升了风险感知状态抽象、连续控制建模和长时序动力学预测。
% 在多种仿真驾驶场景上的大量实验表明，LIDAR-AD 持续优于世界模型基线，并在基于学习的方法中取得最高回报和最佳成功率。
% 在 nuPlan 衍生的日志重构场景上的评估进一步证明了 LIDAR-AD 在真实交通布局下的可迁移性。
% 项目页面见 \href{https://lumen2023.github.io/lidar-ad-website/}{Project Website}。
%
\begin{abstract}
Autonomous driving requires long-horizon closed-loop decision making in dynamic traffic environments.
Latent world models offer an effective framework for this problem by enabling imagination-based decision making in compact latent spaces.
However, multi-source observations contain control-irrelevant redundancy, whereas reliable driving decisions rely on risk-relevant relations, future dynamics, and continuous action adjustments.
This mismatch makes observation reconstruction and absolute action modeling suboptimal for learning decision-relevant latent dynamics.
We propose LIDAR-AD, a decoder-free Latent-Interaction Dreamer with Action-Residual Chains for autonomous driving.
LIDAR-AD replaces observation reconstruction with redundancy-reduced latent alignment, encouraging compact representations of risk-relevant relations in multi-source driving inputs.
It further models vehicle control as residual action updates and uses residual-action sequence contrastive learning to align multi-step residual-driven rollouts with future latent states.
A deterministic analysis shows that the latent-tanh residual parameterization preserves interior action reachability while representing smooth long-horizon control as compact local updates.
Together, these designs improve risk-aware state abstraction, continuous-control modeling, and long-horizon dynamics prediction.
Extensive experiments across diverse simulated driving scenarios demonstrate that LIDAR-AD consistently outperforms world-model baselines, achieving the highest reward and the best success rate among learning-based methods.
Evaluations on nuPlan-derived log-reconstructed scenarios further demonstrate the transferability of LIDAR-AD under real-world traffic layouts.
The project page is available at \href{https://lumen2023.github.io/lidar-ad-website/}{Project Website}.
\end{abstract}

% ===========================================================================
% 关键词 (IEEE Keywords)
% ===========================================================================
\begin{IEEEkeywords}
Autonomous driving, world model, reinforcement learning, 
latent representation, residual actions.
\end{IEEEkeywords}

% ===========================================================================
% 正文 — 从独立章节文件引入
% ===========================================================================
% 中文注释：各章节按逻辑顺序引入。编译时交叉引用和参考文献可在各章节间正常解析。
% 写作时可在对应 .tex 文件中独立编辑，通过编译 main.tex 查看整体效果。
\section{Introduction}
\label{sec:intro}

% 写作指导：自动驾驶是长序列闭环决策，非孤立单步控制。
\IEEEPARstart{A}{utonomous} driving requires reliable decisions in dynamic, interactive, and partially observable traffic environments~\cite{Paden2016Survey}.
Unlike standard continuous-control benchmarks, it is a long-horizon closed-loop problem governed by safety, efficiency, and comfort.
A driving policy must reason about surrounding traffic, maintain stable vehicle-level control, and anticipate the multi-step consequences of current decisions.
Small errors in state representation, action smoothness, or dynamics prediction can therefore accumulate into unsafe or inefficient behavior in dense traffic, intersections, and roundabouts.
Thus, closed-loop driving requires models that extract decision-relevant structure from complex observations while preserving smooth and predictive control.
% 中文翻译：自动驾驶要求系统在动态、交互且部分可观测的交通环境中做出可靠决策。
% 与标准连续控制基准不同，自动驾驶是在安全性、效率和舒适性约束下的长时序闭环问题。
% 驾驶策略必须推理周围交通状态，保持稳定的车辆级控制，并预测当前决策的多步后果。
% 因此，在高密度交通、交叉路口和环岛等场景中，状态表征、动作平滑性或动力学预测中的小误差都可能累积为不安全或低效行为。
% 因而，闭环驾驶需要模型从复杂观测中提取与决策相关的结构，同时保持平滑且具预测性的控制。

\begin{figure}[t]
\centering
\includegraphics[width=0.95\linewidth]{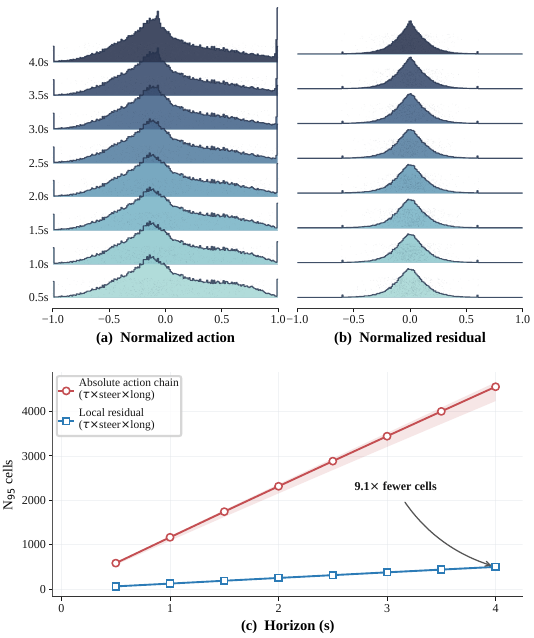}
\caption{Compactness of residual-action space in real driving data.
Statistics from NAVSIM and nuPlan show that consecutive steering and longitudinal commands are strongly correlated, while their residual changes are concentrated in a much narrower range.
The residual-action horizon occupies 9.1$\times$ fewer cells than the absolute-action chain, indicating a more compact geometric structure for smooth long-horizon driving control.}
\label{fig:residual_action_horizon}
\vspace{-10pt}
\end{figure}
% 中文翻译：真实驾驶数据中的残差动作空间紧凑性。
% NAVSIM 和 nuPlan 的统计结果表明，连续转向和纵向控制指令高度相关，而它们的残差变化集中在更窄的范围内。
% 残差动作 horizon 相比绝对动作链占用的网格数量减少 9.1 倍，说明其对平滑长时序驾驶控制具有更紧凑的几何结构。

% 写作指导：将 related work 融合到 introduction 的主线中，避免独立综述式铺陈。
Reinforcement learning has been widely studied for autonomous driving, but model-free policies still require extensive environment interaction, which is costly for rare safety-critical events~\cite{Kiran2022Deep}.
Model-based reinforcement learning alleviates this issue by learning predictive world models that reuse collected experience for imagined rollouts, from short-rollout model learning to RSSM-based latent imagination and value-aware latent planning~\cite{Janner2019MBPO,Hafner2019PlaNet,Hafner2023DreamerV3,Hansen2022TDMPC,Hansen2024TDMPC2}.
In autonomous driving, world models have been adapted through geometric priors, semantic abstraction, safety-aware optimization, risk-aware policy coordination, decoder-free training, and generative scene evolution~\cite{Hu2022MILE,Gao2024SEM2,Huang2024SafeDreamer,Liu2025ITSC,Morihira2026R2Dreamer,Li2024Think2Drive,Wang2023DriveDreamer,Zheng2024GenAD}.
These studies demonstrate the value of learned latent dynamics for prediction, planning, and decision making.
However, two connected issues remain insufficiently addressed for closed-loop driving.
First, many world models still allocate capacity to redundant or weakly decision-relevant observations, rather than organizing latent states around interaction changes that directly affect driving decisions.
Second, conventional action conditioning treats each action as an independent input, offering little structure for continuous steering and longitudinal control.
As a result, the model may fail to associate small control adjustments with future state transitions, while the policy may produce unnecessary action fluctuations during closed-loop execution.

% 中文翻译：强化学习已被广泛用于自动驾驶研究，但免模型策略仍然需要大量环境交互，而稀缺的安全关键事件采集成本高。
% 基于模型的强化学习通过学习预测性世界模型缓解这一问题，相关方法从短 rollout 模型学习扩展到 RSSM 潜在想象和值感知潜在规划。
% 在自动驾驶中，世界模型通过几何先验、语义抽象、安全感知优化、风险感知策略协调、无解码器训练和生成式场景演化等方式被进一步适配到驾驶任务。
% 这些研究证明了学习型潜在动力学在预测、规划和决策中的价值。
% 然而，对于闭环驾驶而言，仍有两个相互关联的问题没有得到充分解决。
% 第一，许多世界模型仍然将建模容量分配给冗余或与决策弱相关的观测信息，而不是围绕直接影响驾驶决策的交互变化来组织潜在状态。
% 第二，传统动作条件化将每个动作视为独立输入，缺少对连续转向和纵向控制的结构表达。
% 因此，模型可能难以将小幅控制调整与未来状态转移联系起来，策略也可能在闭环执行中产生不必要的动作波动。

% 写作指导：讲冗余观测建模导致表征容量浪费，并纳入 decoder-free related work。
Observation reconstruction provides dense supervision for world model learning, but it can bias the latent state toward predictable regularities that are weakly related to control.
Decoder-free representation learning addresses this limitation by optimizing latent embeddings directly through predictive, contrastive, redundancy-reduction, and joint-embedding objectives~\cite{Oord2018CPC,Grill2020BYOL,Zbontar2021Barlow,Assran2023IJEPA,Bardes2024VJEPA}.
These objectives have also improved reinforcement learning and latent dynamics modeling through contrastive, predictive, or augmentation-based visual control~\cite{Laskin2020CURL,Schwarzer2021SPR,Kostrikov2021DrQ,Yarats2022DrQv2}.
Nevertheless, most decoder-free objectives are designed for holistic inputs or augmented views, and provide limited mechanisms for modeling structured relationships among heterogeneous driving modalities.
In driving sequences, many observed factors remain temporally persistent, whereas safety-critical changes often appear sparsely around interaction boundaries, such as abrupt range variations, obstacle encounters, yielding, merging, or entering a conflict region.
When the objective emphasizes generic observation prediction, the model may reduce loss by modeling persistence rather than identifying interaction changes that drive future decisions.
This can consume modeling capacity and weaken long-horizon closed-loop decision making.
% 中文翻译：观测重建为世界模型学习提供了稠密信号，但也可能使潜在状态偏向于容易建模却与控制弱相关的观测规律。
% 无解码器表征学习通过预测式、对比式、冗余消减和联合嵌入目标直接优化潜在嵌入，从而缓解这一限制。
% 这些目标也通过对比、预测或数据增强式视觉控制改进了强化学习和潜在动力学建模。
% 然而，大多数无解码器目标主要面向整体输入或增强视图，对异构驾驶模态之间结构化关系的建模机制仍然有限。
% 在驾驶序列中，许多观测因素具有较强时间持续性，而安全关键变化往往稀疏地出现在交互边界附近，
% 例如突发距离变化、障碍物出现、让行、汇入或进入冲突区域。
% 当学习目标强调通用观测预测时，模型可能通过建模时间持续性来降低损失，而不是识别真正驱动未来决策的交互变化。
% 这会消耗模型容量，并削弱长时序闭环决策能力。

% 写作指导：闭环驾驶应看作围绕上一控制命令的连续修正，非每步重选绝对动作。
Another limitation lies in how closed-loop driving actions are commonly represented.
Many learning-based policies output a new control command at each decision step, which gives limited structure to the temporal continuity of steering and longitudinal control.
Residual reinforcement learning and smooth-control regularization suggest that learning local corrections or suppressing abrupt action changes can improve continuous-control stability~\cite{Johannink2019ResidualRL,Mysore2021CAPS}.
Recent residual-action methods further use the previous action as the reference signal, reducing the effective search space and improving long-horizon prediction or action smoothness in visual control~\cite{Liu2025ResAct,Zhang2026ResWM}.
In real driving, maneuvers such as lane changes, controlled stops, and collision avoidance are rarely produced by isolated commands; instead, they are formed by gradual adjustments around previous controls.
As shown in Fig.~\ref{fig:residual_action_horizon}, our statistics on NAVSIM and nuPlan show that consecutive driving commands are highly correlated, and their residual changes occupy a much smaller range than the full action space.
This suggests that repeatedly predicting absolute actions may introduce unnecessary modeling burden, especially when temporally redundant control patterns persist over many steps.
However, a single residual action describes only a local correction.
Long-horizon driving instead requires modeling how intermediate control changes accumulate into future outcomes, an abstraction also reflected in sequence-level action generation and chain-style action modeling~\cite{Chi2023DiffusionPolicy,Zhang2025CoA}.
Thus, the world model should learn how successive per-step residual-action inputs shape future driving states.
% 中文翻译：另一个限制在于闭环驾驶动作通常如何被表示。
% 许多基于学习的驾驶策略在每个决策步输出一个新的控制命令，这对转向和纵向控制的时间连续性提供的结构有限。
% 残差强化学习和平滑控制正则化表明，学习局部修正或抑制动作突变有助于提升连续控制稳定性。
% 近期残差动作方法进一步以前一动作作为参考信号，从而缩小有效搜索空间，并改善视觉控制中的长时序预测或动作平滑性。
% 在真实驾驶中，变道、平稳停车和避障等机动很少由孤立命令产生，而通常是在上一控制命令基础上逐渐调整形成的。
% 如图~\ref{fig:residual_action_horizon} 所示，我们在 NAVSIM 和 nuPlan 上的统计结果表明，连续驾驶控制命令高度相关，
% 它们的残差变化范围远小于完整动作空间。
% 这说明反复预测绝对动作可能引入不必要的建模负担，尤其是当时间上冗余的控制模式会持续多个步骤时。
% 然而，单个残差动作仍然只能描述局部修正。
% 长时序驾驶需要建模中间控制变化如何累积形成未来结果，这一抽象也体现在序列级动作生成和链式动作建模中。
% 因此，世界模型还应学习逐步 residual-action 输入如何塑造未来驾驶状态。

Table~\ref{tab:method_comparison} summarizes the methodological position of LIDAR-AD relative to representative decoder-free, residual-action, and world-model baselines.
In contrast to methods that address representation compactness or residual control separately, LIDAR-AD combines decoder-free representation learning, group-level observation interaction, residual-action policy parameterization, residual-aware transition modeling, and multi-step action contrast within one RSSM-based framework.

% 写作指导：简要引出方法，只回应前文问题，不展开模块细节。
To address these challenges, this paper proposes \textbf{LIDAR-AD}, a decoder-free latent-interaction Dreamer with action-residual chains for autonomous driving.
The central idea is to guide the world model toward decision-relevant interaction changes and continuous control evolution, rather than treating observation modeling and action prediction as separate one-step objectives.
LIDAR-AD integrates \textbf{Decoder-Free Latent Interaction Representation (DLIR)}, a \textbf{Residual-Action World Model (RAWM)}, and \textbf{Residual-Action Chain Contrastive Learning (ARC-CL)} within a unified RSSM framework.
This design reduces redundant observation modeling, provides a compact action space for smooth control, and aligns multi-step residual-action rollouts with future latent states.
% 中文翻译：为解决上述挑战，本文提出 LIDAR-AD，一种面向自动驾驶的无解码器潜在交互 Dreamer，其核心是动作残差链建模。
% 其核心思想是引导世界模型关注与决策相关的交互变化和连续控制演化，而不是将观测建模和动作预测视为相互分离的一步目标。
% LIDAR-AD 在统一的 RSSM 框架中整合了无解码器潜在交互表征（DLIR）、残差动作世界模型（RAWM）和残差动作链对比学习（ARC-CL）。
% 该设计减少冗余观测建模，为平滑控制提供紧凑动作空间，并将多步 residual-action rollout 与未来潜在状态对齐。

% 写作指导：贡献列表只写三个主要方法贡献。
The main contributions of this paper are summarized as follows:
\begin{itemize}[leftmargin=*, itemsep=0.8ex]
\item \textbf{Decoder-Free Latent Interaction Representation (DLIR).}
We develop a lightweight representation learner that models interactions among ego state, range sensing, and navigation signals, and aligns the RSSM latent state with an interaction-aware embedding without reconstructing the full observation.

\item \textbf{Residual-Action World Model (RAWM).}
We introduce a residual-action parameterization into an RSSM-based world model, where the transition is conditioned on both the executed action and the residual adjustment.
A deterministic analysis characterizes command invariance, interior reachability, and local compactness under a smooth-control assumption.

\item \textbf{Residual-Action Chain Contrastive Learning (ARC-CL).}
We design a multi-step contrastive objective that aligns rollouts recursively driven by per-step residual-action inputs with true future latent states, improving the world model's ability to capture cumulative control consequences.
\end{itemize}
% 中文翻译：本文主要贡献包括：
% （1）无解码器潜在交互表征（DLIR）。本文提出一种轻量级表征学习器，建模自车状态、距离感知和导航信号之间的交互，并在不重建完整观测的情况下，
% 将 RSSM 潜在状态与交互感知嵌入对齐。
% （2）残差动作世界模型（RAWM）。本文将残差动作参数化引入基于 RSSM 的世界模型，
% 使转移模型同时条件化于执行动作和残差调整。
% 确定性分析刻画命令不变性、内部可达性，以及平滑控制假设下的局部紧凑性。
% （3）残差动作链对比学习（ARC-CL）。本文设计一种多步对比目标，将由逐步 residual-action 输入递归驱动的 rollout 与真实未来潜在状态对齐，
% 提升世界模型捕获累积控制后果的能力。

% 写作指导：结尾只做结构导航，保持简洁。
The remainder of this paper is organized as follows.
Section~\ref{sec:problem} formulates the driving problem and defines the observation, action, and latent state spaces.
Section~\ref{sec:lidarad} presents the LIDAR-AD framework.
Section~\ref{sec:exp_setup} describes the experimental setup, and Section~\ref{sec:exp_results} reports the results.
Section~\ref{sec:discussion} discusses implications and limitations, and Section~\ref{sec:conclusion} concludes the paper.
% 中文翻译：本文其余部分组织如下。
% 后续章节依次形式化驾驶问题，介绍 LIDAR-AD 框架，描述实验设置并报告结果，最后讨论启示、局限并总结全文。

\begin{table*}[t]
  \centering
  \caption{Component-Level Comparison with Related Methods}
  \label{tab:method_comparison}
  \setlength{\tabcolsep}{7pt}
  \renewcommand{\arraystretch}{1.12}
  \footnotesize
  \begin{tabular}{lccccc}
    \toprule
    \textbf{Method} &
    \textbf{Decoder-Free} &
    \textbf{Group Interaction} &
    \textbf{Residual Actor} &
    \textbf{Residual-Aware Transition} &
    \textbf{Multi-Step Action Contrast} \\
    \midrule
    DreamerV3~\cite{Hafner2023DreamerV3} & \CNo & \CNo & \CNo & \CNo & \CNo \\
    R2-Dreamer~\cite{Morihira2026R2Dreamer} & \CYes & \CNo & \CNo & \CNo & \CNo \\
    ResAct~\cite{Liu2025ResAct} & \CNo & \CNo & \CYes & \CNo & \CNo \\
    ResWM~\cite{Zhang2026ResWM} & \CNo & \CNo & \CYes & \CYes & \CNo \\
    \textbf{LIDAR-AD(Ours)} & \CYes & \CYes & \CYes & \CYes & \CYes \\
    \bottomrule
  \end{tabular}

  \vspace{0.5\baselineskip}
  \parbox{0.96\textwidth}{
  \footnotesize
  \emph{Note:} \CYes{} indicates that the method explicitly includes the corresponding component, while \CNo{} indicates that the component is not part of the original method design.
  }
\end{table*}

\section{Problem Formulation}
\label{sec:problem}

% 写作引导：按任务定义→世界模型→残差动作顺序组织，聚焦 latent dynamics 核心

\subsection{Reinforcement Learning for Autonomous Driving}
\label{sec:problem_rl}

We formulate autonomous driving as a partially observable sequential decision-making problem.
At each timestep $t$, the agent receives an observation $o_t$, executes a continuous control action $a_t$, and obtains a reward $\rho_t$ from the environment.
The task is modeled as a partially observable Markov decision process (POMDP):
\begin{equation}
\mathcal{M} = (\mathcal{O}, \mathcal{A}, P, \mathcal{R}, \gamma),
\label{eq:pomdp_tuple}
\end{equation}
where $\mathcal{O}$ is the observation space, $\mathcal{A}$ is the continuous action space, $P(o_{t+1} \mid o_{\leq t}, a_t)$ denotes the unknown history-conditioned transition dynamics, $\mathcal{R}(o_t,a_t)$ is the reward function, and $\gamma \in [0,1)$ is the discount factor.
The policy $\pi(a_t \mid o_{\leq t})$ is optimized to maximize the expected discounted return:
\begin{equation}
\max_\pi \; \mathbb{E}_{\pi} \left[
\sum_{t=0}^{\infty} \gamma^t \rho_t
\right].
\label{eq:rl_objective}
\end{equation}
Here $\rho_t=\mathcal{R}(o_t,a_t)$ denotes the realized scalar reward at timestep $t$.
In our driving setting, $a_t$ denotes the executed continuous control command, including longitudinal and lateral control such as acceleration and steering.
The concrete observation composition, action range, reward design, and simulator configuration are reported in Section~\ref{sec:exp_setup}.
% 中文翻译：
% 本文将自动驾驶形式化为部分可观测的序列决策问题。
% 在每个时间步 $t$，智能体接收观测 $o_t$，执行连续控制动作 $a_t$，并从环境获得奖励 $\rho_t$。
% 该任务被建模为 POMDP：$\mathcal{M}=(\mathcal{O},\mathcal{A},P,\mathcal{R},\gamma)$。
% 其中，$\mathcal{O}$ 是观测空间，$\mathcal{A}$ 是连续动作空间，
% $P(o_{t+1}\mid o_{\leq t},a_t)$ 表示未知的历史条件转移动力学，
% $\mathcal{R}(o_t,a_t)$ 是奖励函数，$\gamma\in[0,1)$ 是折扣因子。
% 策略 $\pi(a_t\mid o_{\leq t})$ 通过最大化期望折扣回报进行优化。
% 这里 $\rho_t=\mathcal{R}(o_t,a_t)$ 表示时间步 $t$ 的实际标量奖励。
% 在本文驾驶设置中，$a_t$ 表示实际执行的连续控制命令，包括加速度和转向等纵向与横向控制。
% 具体的观测组成、动作范围、奖励设计和仿真器配置见第~\ref{sec:exp_setup} 节。

% 写作引导：说明 RSSM 为世界模型基础，只写 latent state/prior-posterior 与 policy 关系

\noindent\textbf{RSSM latent state.}
Following the reconstruction-free RSSM formulation in R2-Dreamer~\cite{Morihira2026R2Dreamer}, we learn compact latent dynamics for policy optimization without reconstructing raw observations.
The observation history is summarized into a latent state
\begin{equation}
s_t = (h_t,z_t),
\label{eq:latent_state}
\end{equation}
where $h_t$ is a deterministic recurrent state and $z_t$ is a stochastic latent state.
The recurrent transition summarizes historical information as
\begin{equation}
h_t = f_\phi(h_{t-1}, z_{t-1}, \xi_{t-1}),
\label{eq:rssm_transition}
\end{equation}
while the stochastic state is inferred from the current observation embedding or predicted from the prior:
\begin{equation}
z_t \sim q_\phi(z_t \mid h_t, e_t),
\qquad
\hat{z}_t \sim p_\phi(\hat{z}_t \mid h_t),
\label{eq:rssm_posterior_prior}
\end{equation}
where $e_t$ denotes the latent embedding of the current observation.
The transition input $\xi_{t-1}$ is the executed action $a_{t-1}$ in a conventional action-conditioned RSSM; in RAWM, it is replaced by the residual-action embedding $d_{t-1}$ defined in Section~\ref{sec:problem_residual_chain}.
The learned latent state supports reward prediction, continuation prediction, imagined rollout, and actor-critic policy learning.
Different from reconstruction-based world models, the representation is optimized in latent space, reducing the need to supervise every raw observation coordinate and encouraging compact task-relevant dynamics.
% 中文翻译：
% 遵循 R2-Dreamer 中的无重建 RSSM 形式，本文在不重建原始观测的情况下学习用于策略优化的紧凑潜在动力学。
% 观测历史被概括为潜在状态 $s_t=(h_t,z_t)$，其中 $h_t$ 是确定性循环状态，$z_t$ 是随机潜在状态。
% 循环转移 $h_t=f_\phi(h_{t-1},z_{t-1},\xi_{t-1})$ 用于汇总历史信息。
% 随机状态可以由当前观测嵌入推断得到，也可以由先验预测得到，即
% $z_t\sim q_\phi(z_t\mid h_t,e_t)$，$\hat{z}_t\sim p_\phi(\hat{z}_t\mid h_t)$。
% 其中 $e_t$ 表示当前观测的潜在嵌入。
% 在传统动作条件 RSSM 中，转移输入 $\xi_{t-1}$ 是执行动作 $a_{t-1}$；
% 在 RAWM 中，它被第~\ref{sec:problem_residual_chain} 节定义的残差动作嵌入 $d_{t-1}$ 替代。
% 学到的潜在状态支持奖励预测、continue 预测、想象 rollout 和 actor-critic 策略学习。
% 与基于重建的世界模型不同，该表征直接在潜在空间中优化，
% 从而减少对每个原始观测坐标的监督需求，并鼓励紧凑的任务相关动力学。

\begin{figure}[t]
\centering
\includegraphics[width=\linewidth]{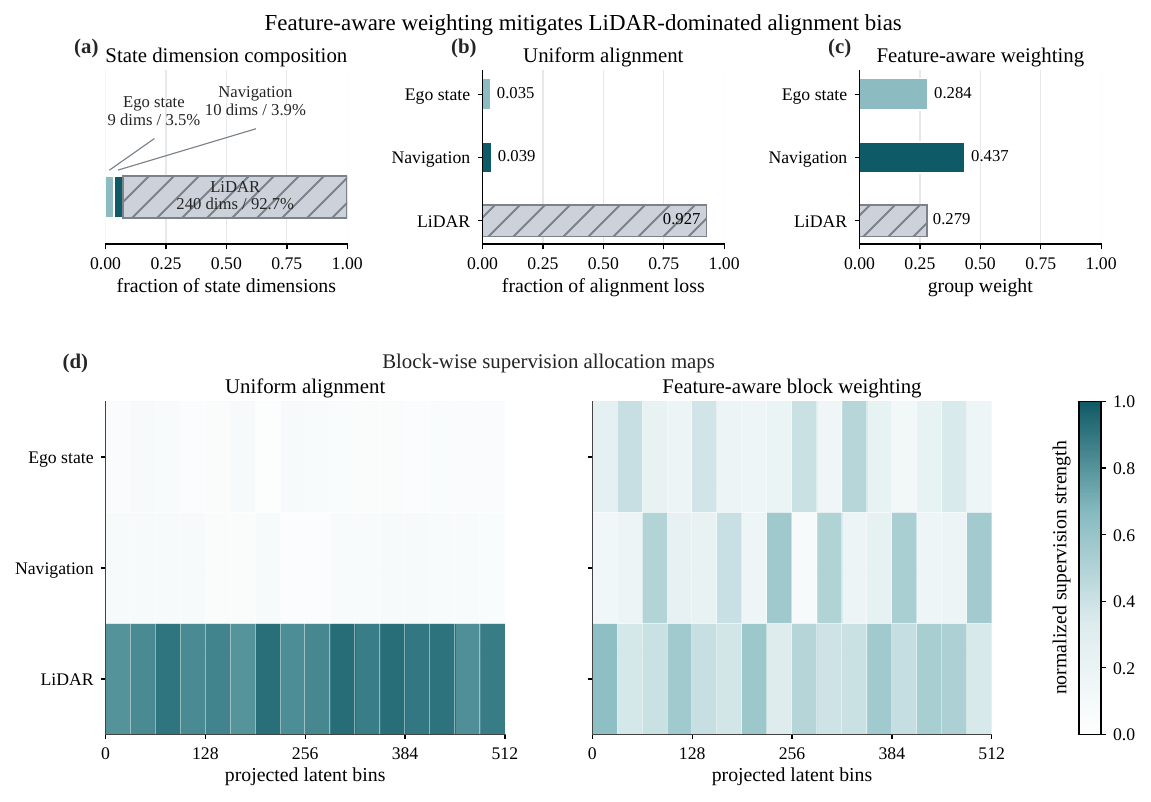}
\caption{Feature-aware supervision allocation for multi-source latent alignment.
Uniform alignment is dominated by high-dimensional LiDAR inputs, 
while feature-aware weighting redistributes 
supervision across ego-state, navigation, and LiDAR-related latent bins.}
\label{fig:feature_alignment_bias}
\vspace{-4pt}
\end{figure}
% 图片注释中文翻译：
% 潜在动力学中的模态尺寸诱导监督偏置。
% 对异构驾驶观测，均匀监督下高维 LiDAR 特征可能贡献更大的总体对齐信号，
% 从而弱化紧凑风险或交互线索，因而需要 DLIR 中的风险加权潜在对齐。

Fig.~\ref{fig:feature_alignment_bias} illustrates this modality-size-induced supervision bias.
It shows why a uniform alignment objective can become LiDAR-dominated when heterogeneous modalities have substantially different input dimensions, and motivates balancing the latent supervision across compact risk and interaction cues in DLIR.
% 中文翻译：
% 图~\ref{fig:feature_alignment_bias} 展示了这种由模态尺寸诱导的监督偏置。
% 当异构模态具有显著不同的输入维度时，均匀对齐目标可能被高维 LiDAR 特征主导。
% 这进一步说明，在 DLIR 中需要在紧凑风险线索和交互线索之间平衡潜在监督。

In this work, the RSSM backbone is not redesigned.
Instead, we focus on how to make its latent dynamics more suitable for autonomous driving.
Specifically, the observation embedding $e_t$ should preserve decision-relevant interactions among heterogeneous driving cues, and the transition model should account for the temporal continuity of executed control commands.
The former is addressed by the proposed latent interaction representation, while the latter motivates residual-action-aware world modeling and multi-step action-sequence consistency learning.
% 中文翻译：
% 本文不重新设计 RSSM 主干。
% 相反，本文关注如何使其潜在动力学更适合自动驾驶任务。
% 具体而言，观测嵌入 $e_t$ 应保留异构驾驶线索之间与决策相关的交互，
% 转移模型则应刻画已执行控制命令的时间连续性。
% 前者由本文提出的潜在交互表征处理，后者则引出残差动作感知世界模型和多步动作序列一致性学习。

\subsection{Residual Actions and Multi-Step Alignment Target}
\label{sec:problem_residual_chain}

Driving control is inherently continuous over time.
We therefore formulate residual control as an explicit modeling variable, rather than treating it as a post-hoc difference between consecutive executed actions.
Let $a_{t-1}\in(-1,1)^d$ denote the previously executed action, and let $\Delta u_t\in\mathbb{R}^d$ be the residual command predicted by the actor at timestep $t$.
The residual is defined in the pre-tanh action space by first mapping the previous executed action back to its unconstrained coordinate:

% 中文翻译：
% 驾驶控制在时间上具有内在连续性。
% 因此，本文将残差控制形式化为显式建模变量，而不是将其视为连续执行动作之间的事后差分。
% 令 $a_{t-1}\in(-1,1)^d$ 表示上一时刻已执行动作，
% $\Delta u_t\in\mathbb{R}^d$ 表示 actor 在时间步 $t$ 预测的残差命令。
% 残差定义在 tanh 之前的动作空间中；具体做法是先将上一已执行动作映射回无约束坐标。
% 核心逻辑：residual 是 actor 显式输出的控制调整，定义在 pre-tanh 空间。

\begin{equation}
u_{t-1}  =
\operatorname{atanh}
\left(
\operatorname{clip}(a_{t-1}, -1+\epsilon, 1-\epsilon)
\right),
\label{eq:prev_action_latent}
\end{equation}
where $\epsilon$ is a small constant introduced for numerical stability.
The current executed action is then obtained as
\begin{equation}
a_t =
\tanh\left(u_{t-1}+\Delta u_t\right).
\label{eq:latent_tanh_action}
\end{equation}
Here, $\Delta u_t$ denotes the actor-produced residual control command, whereas $a_t$ is the bounded action applied to the environment.

% 中文翻译：
% 其中 $\epsilon$ 是为数值稳定性引入的小常数。
% 当前执行动作通过 $a_t=\tanh(u_{t-1}+\Delta u_t)$ 得到。
% 这里，$\Delta u_t$ 表示由 actor 产生的残差控制命令，而 $a_t$ 是实际施加到环境中的有界动作。
% 核心逻辑：策略学习局部控制调整，而不是在每一步重新生成完整绝对动作。

To describe the temporal evolution of recent control adjustments, we use an ordered residual-action sequence over a short horizon $K$:
\begin{equation}
\mathcal{Q}_{t}^{K} =
\left\{(a_{t+k},\Delta u_{t+k})\right\}_{k=0}^{K-1},
\label{eq:residual_action_chain}
\end{equation}
where $K$ denotes the rollout horizon.
This notation does not introduce a standalone feature vector for the whole chain.
It only identifies the per-step executed actions and residual commands used by ARC-CL to supervise a multi-step prior rollout.

% 中文翻译：
% 为描述近期控制调整的时间演化，本文使用短时域 $K$ 上的有序残差动作序列 $\mathcal{Q}_t^K$。
% 其中 $K$ 表示 rollout 时域。
% 该记号并不引入一个表示整条动作链的独立特征向量。
% 它只标识 ARC-CL 用于监督多步先验 rollout 的逐步执行动作和残差命令。
% 核心逻辑：ARC-CL 关注连续残差序列的累积后果，但不额外构造整体链向量。

At each transition step, the latent world model receives the executed action and the residual command at that same step:
\begin{equation}
d_t = E_{\mathrm{act}}([a_t;\Delta u_t]),
\qquad
s_{t+1}^{p}=T_\phi(s_t,d_t).
\label{eq:per_step_residual_transition}
\end{equation}
Thus, the RSSM transition is conditioned on per-step joint action information, not on an encoded action-chain vector.
ARC-CL later aligns the $K$-step rollout obtained by repeatedly applying this transition over $\mathcal{Q}_{t}^{K}$ with the future posterior state.
The detailed transition parameterization and structural properties are presented in Section~\ref{sec:rawm} and Section~\ref{sec:residual_theory}.

% 中文翻译：
% 在每个转移步中，潜在世界模型接收同一时间步的执行动作和残差命令。
% 动作嵌入由 $d_t=E_{\mathrm{act}}([a_t;\Delta u_t])$ 给出，
% 先验潜在状态由 $s_{t+1}^{p}=T_\phi(s_t,d_t)$ 转移得到。
% 因此，RSSM 转移条件化于逐步联合动作信息，而不是条件化于编码后的动作链向量。
% ARC-CL 随后将通过在 $\mathcal{Q}_t^K$ 上重复应用该转移得到的 $K$ 步 rollout 与未来后验状态对齐。
% 详细的转移参数化和结构性质见第~\ref{sec:rawm} 节和第~\ref{sec:residual_theory} 节。
% 核心逻辑：RAWM 输入逐步 $[a_t;\Delta u_t]$，ARC-CL 执行多步对齐，但不使用整体链向量。

\begin{figure*}[t]
\centering
{\pdfimageresolution=300
\includegraphics[width=1.00\textwidth]{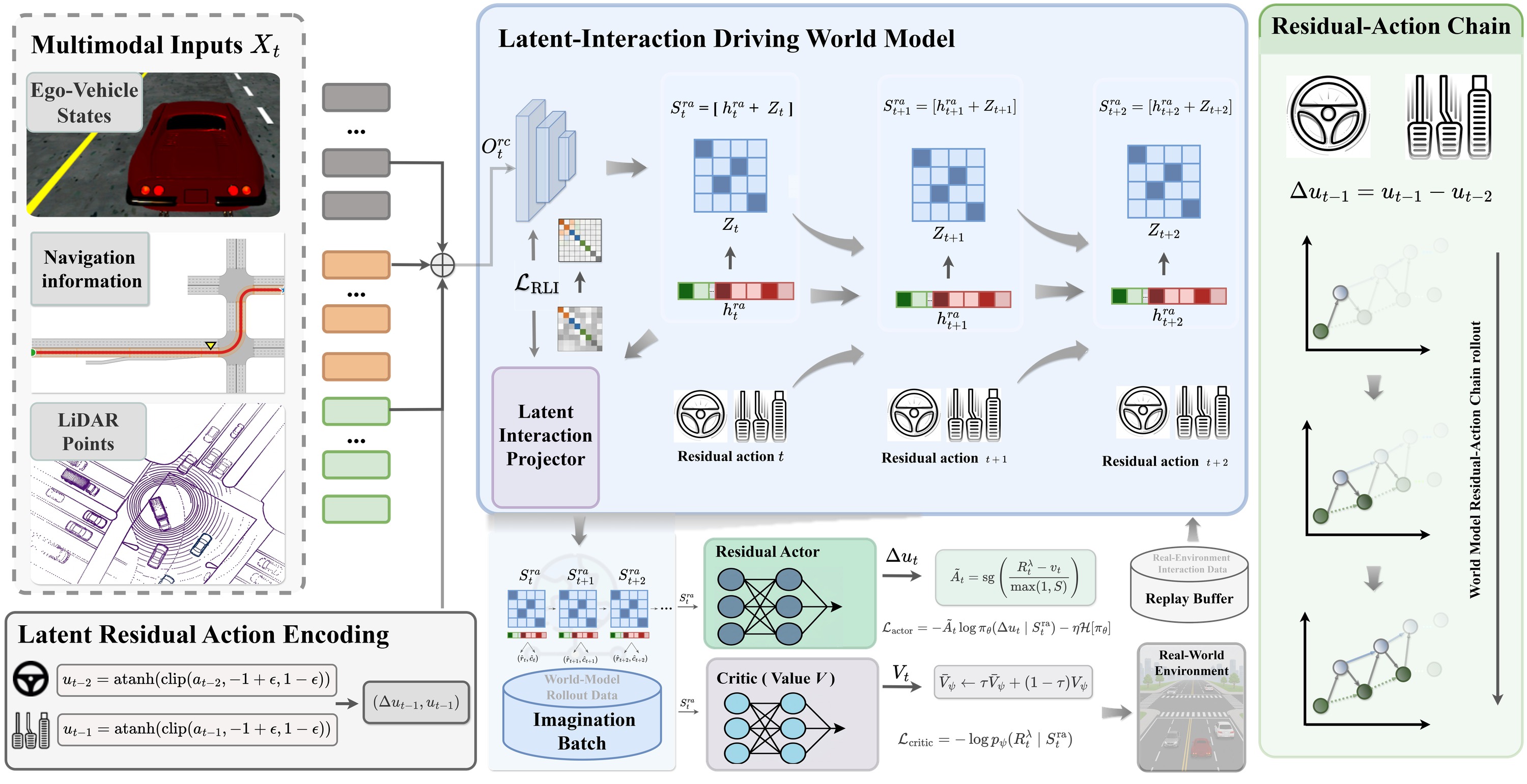}
}
\caption{Overall architecture of LIDAR-AD.
DLIR maps heterogeneous driving observations to the shared RSSM latent state, RAWM converts actor residuals into executable actions and per-step action-change embeddings, and ARC-CL aligns recursive multi-step prior rollouts with future posterior states.}
\label{fig:lidarad_framework}
\vspace{-4pt}
\end{figure*}
% 图片注释中文翻译：LIDAR-AD整体架构：DLIR映射异构观测到共享RSSM潜在状态，RAWM转换actor残差为可执行动作和逐步动作变化嵌入，ARC-CL对齐递归多步rollout与未来后验。

\section{Method}
\label{sec:lidarad}

LIDAR-AD is a decoder-free latent world-model framework for closed-loop autonomous driving.
It uses the RSSM state $s_t=(h_t,z_t)$ as a common latent interface for observation abstraction, residual control, and multi-step dynamics learning.
Fig.~\ref{fig:lidarad_framework} shows the architecture, and the following subsections define how DLIR, RAWM, and ARC-CL constrain the shared latent dynamics.

% 中文翻译：
% LIDAR-AD 是一种面向闭环自动驾驶的无解码器潜在世界模型框架。
% 该框架使用 RSSM 状态 $s_t=(h_t,z_t)$ 作为观测抽象、残差控制和多步动力学学习之间的公共潜在接口。
% 图~\ref{fig:lidarad_framework} 给出了整体架构，后续小节定义 DLIR、RAWM 和 ARC-CL 如何约束这一共享潜在动力学。

\subsection{Framework Overview}
\label{sec:overview}

At each timestep $t$, the agent receives the complete driving observation $o_t=\{x_t^m\}_{m\in\mathcal{M}}$, where $x_t^m$ denotes the $m$-th semantic observation group and $\mathcal{M}$ is the set of observation groups.
The framework imposes three complementary constraints on the shared RSSM dynamics: DLIR selects interaction-relevant observation content, RAWM specifies how residual actions enter the transition, and ARC-CL aligns multi-step action effects.
DLIR encodes $o_t$ into $e_t^{\mathrm{DLIR}}$, from which the RSSM posterior infers $s_t^q=(h_t,z_t^q)$.
Conditioned on $s_t^q$ and the previous pre-tanh action, the actor predicts $\Delta u_t$; RAWM then decodes $a_t$ and forms the transition input $d_t$.

For multi-step training, ARC-CL recursively applies the same transition over $(d_t,\ldots,d_{t+K-1})$ and aligns the final prior state with the posterior inferred from the future observation.
The ordered embeddings are used only as step-wise transition inputs, not as an additional chain-level feature.
Thus, the overview pipeline is
\[
o_t \rightarrow e_t^{\mathrm{DLIR}} \rightarrow s_t^q
\rightarrow (\Delta u_t,a_t,d_t) \rightarrow s_{t+1}^p,
\]
with ARC-CL extending the last transition over $K$ steps during training.

% 中文翻译：
% 在每个时间步 $t$，智能体接收完整驾驶观测 $o_t=\{x_t^m\}_{m\in\mathcal{M}}$，其中 $x_t^m$ 表示一个语义观测组。
% 该框架从三个互补角度约束共享 RSSM 动力学：DLIR 选择交互相关观测内容，RAWM 定义残差动作如何进入转移模型，ARC-CL 对齐多步动作影响。
% DLIR 将 $o_t$ 编码为 $e_t^{\mathrm{DLIR}}$，RSSM 后验再由该嵌入推断 $s_t^q=(h_t,z_t^q)$。
% 在 $s_t^q$ 和上一 pre-tanh 动作的条件下，actor 预测 $\Delta u_t$；RAWM 将其映射为执行动作 $a_t$，并形成 transition 输入 $d_t$。
% 在多步训练中，ARC-CL 在有序嵌入 $(d_t,\ldots,d_{t+K-1})$ 上递归应用同一个 transition，
% 并将最终先验状态与未来观测推断出的后验状态对齐。
% 这些有序嵌入只作为逐步 transition 输入使用，不会被压缩为额外链级特征。
% 因此，overview 的主流程是
% $o_t \rightarrow e_t^{\mathrm{DLIR}} \rightarrow s_t^q \rightarrow (\Delta u_t,a_t,d_t) \rightarrow s_{t+1}^p$，
% ARC-CL 在训练阶段将最后一步 transition 扩展到 $K$ 步。

% \begin{figure}[!b]
\begin{figure}[t]
\centering
\includegraphics[width=\linewidth]{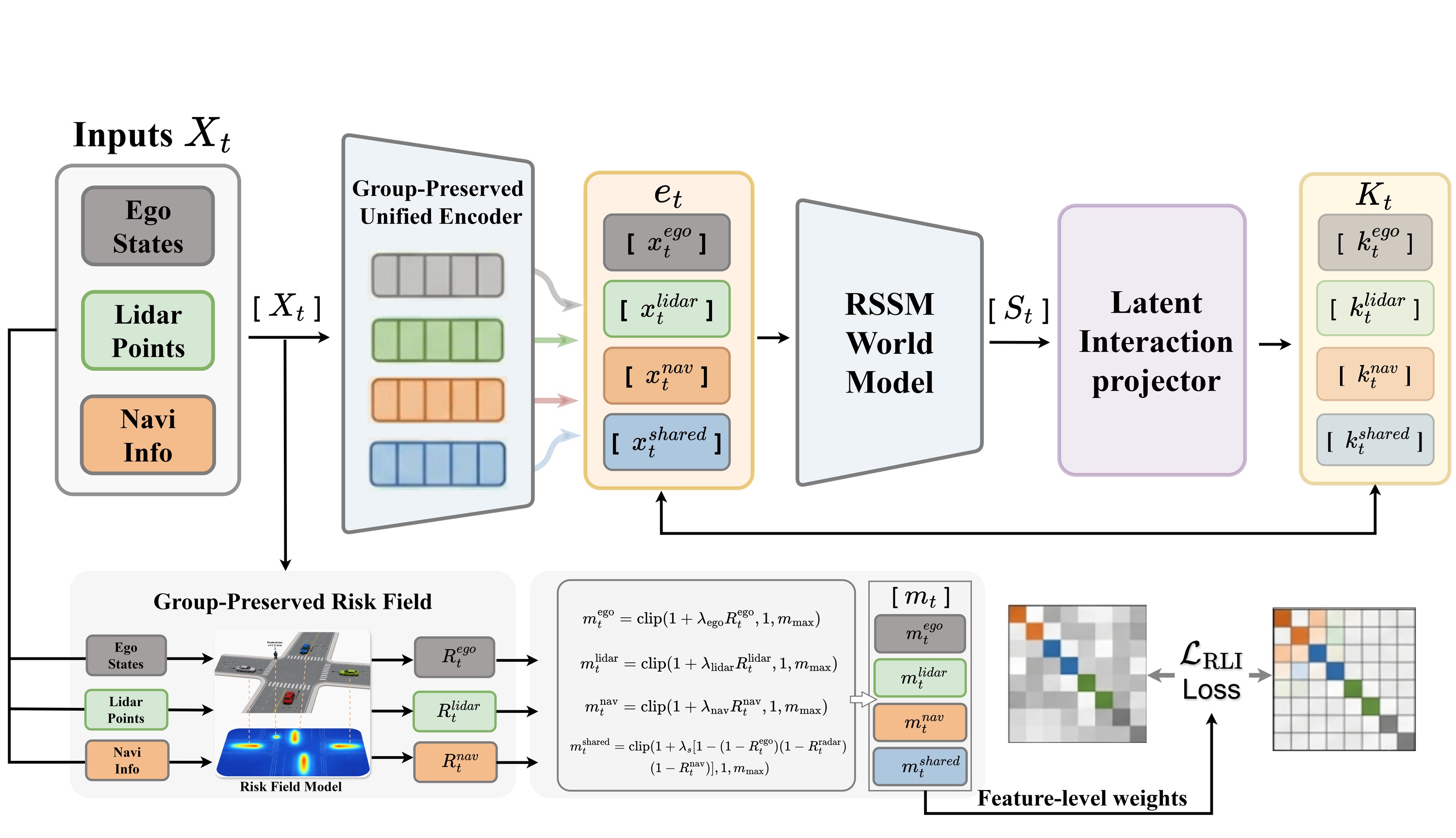}
\caption{Decoder-free latent interaction representation.
Grouped driving observations and the deterministic risk descriptor are encoded separately, fused through gated pairwise interactions, and aligned with the RSSM posterior state by a risk-weighted redundancy-reduction objective without observation reconstruction.}
\label{fig:dlir_framework}
\vspace{-6pt}
\end{figure}
% 图片注释中文翻译：无解码器潜在交互表征：分组观测和确定性风险描述符分别编码，经门控成对交互融合，由风险加权冗余消减目标与RSSM后验对齐，无需观测重建。

\subsection{Decoder-Free Latent Interaction Representation}
\label{sec:dlir}

DLIR replaces observation reconstruction with decoder-free latent alignment for heterogeneous driving observations.
It derives a deterministic risk descriptor from structured observations and fuses it with grouped modality embeddings to form interaction-aware latent features.
The resulting embedding is aligned with the RSSM posterior state by a risk-weighted redundancy-reduction objective inspired by Barlow Twins~\cite{Zbontar2021Barlow}.
For two normalized batch embeddings $Y^A,Y^B\in\mathbb{R}^{B\times D}$, the cross-correlation matrix is
\begin{equation}
C_{ij} = \frac{1}{B}
\sum_{b=1}^{B}
\tilde{Y}^{A}_{b,i}
\tilde{Y}^{B}_{b,j},
\label{eq:bt_corr}
\end{equation}
where $\tilde{Y}$ denotes batch-normalized embeddings.
Unlike the standard equal-weight formulation, DLIR uses the risk embedding to weight latent dimensions, emphasizing interaction-relevant components without decoding the latent state.
% 中文翻译：
% DLIR 使用无解码器潜在对齐目标替代观测重建。
% 它从结构化驾驶观测中确定性计算风险描述符，并将其与分组模态嵌入和成对交互特征融合。
% 得到的嵌入通过风险加权的冗余消减目标与 RSSM 后验状态对齐，该目标借鉴 Barlow Twins 的基本思想。
% 对两个归一化 batch 嵌入 $Y^A,Y^B$，先计算互相关矩阵 $C$；
% 与标准等权重形式不同，DLIR 使用风险嵌入为 latent 维度加权，使交互相关组件获得更强监督，同时不需要将潜在状态解码回观测空间。

\noindent\textbf{Risk-guided grouped encoding.}
Let the driving observation be divided into semantic groups,
\begin{equation}
o_t=\{x_t^m\}_{m\in\mathcal{M}},
\label{eq:grouped_obs}
\end{equation}
where each group corresponds to a different type of driving information.
A deterministic risk descriptor for DLIR is computed as
\begin{equation}
g_t=\Gamma(o_t),
\label{eq:risk_field}
\end{equation}
where $\Gamma(\cdot)$ summarizes proximity and route-consistency cues without introducing an additional supervised prediction target.
In DLIR, $g_t$ is used only to guide decoder-free latent alignment.

\noindent\textbf{Risk descriptor construction.}
We decompose the descriptor into LiDAR, navigation, and ego-state components,
\begin{equation}
g_t=[g_t^{\mathrm{lidar}};g_t^{\mathrm{nav}};g_t^{\mathrm{ego}}],
\label{eq:risk_descriptor_blocks}
\end{equation}
which correspond to environmental proximity, route constraint, and vehicle-state risk, respectively.
For LiDAR-like range sensing with beam distances $\{d_{t,i}\}_{i=1}^{N_l}$, the proximity component is
\begin{equation}
g_t^{\mathrm{lidar}}
=
\left\{
\exp\!\left(-d_{t,i}/\sigma_l\right)
\right\}_{i=1}^{N_l},
\label{eq:lidar_risk}
\end{equation}
where $\sigma_l$ controls the distance decay.
This assigns larger responses to nearby obstacles and suppresses distant returns.
The navigation component is
\begin{equation}
g_t^{\mathrm{nav}}
=
\left[
\min\!\left(1,\frac{|e_t^{\mathrm{lat}}|}{e_{\max}}\right),
\frac{1-\cos(e_t^{\mathrm{head}})}{2},
\min\!\left(1,\frac{|\kappa_t|}{\kappa_{\max}}\right)
\right],
\label{eq:nav_risk}
\end{equation}
where $e_t^{\mathrm{lat}}$ is lateral route error, $e_t^{\mathrm{head}}$ is heading error, and $\kappa_t$ is local route curvature.
When curvature is not directly provided by the navigation interface, it is estimated from successive navigation waypoints as $\kappa_t\approx |\Delta\psi_t^{\mathrm{nav}}|/\Delta s$.
The ego-state component is
\begin{equation}
g_t^{\mathrm{ego}}
=
\left[
\min\!\left(1,\frac{\max(0,v_t-v_{\mathrm{ref}})}{v_{\max}-v_{\mathrm{ref}}+\epsilon}\right),
\frac{|\delta_t|}{\delta_{\max}},
\frac{|a_t^{\mathrm{acc}}|}{a_{\max}}
\right],
\label{eq:ego_risk}
\end{equation}
where $v_t$ is ego speed, $\delta_t$ is steering command, and $a_t^{\mathrm{acc}}$ is longitudinal control.
Before being encoded, each block is normalized as
\begin{equation}
\bar{g}_t^{k}
=
\frac{g_t^{k}-\min(g^{k})}
{\max(g^{k})-\min(g^{k})+\epsilon_g},
\quad k\in\{\mathrm{lidar},\mathrm{nav},\mathrm{ego}\},
\label{eq:risk_norm}
\end{equation}
and the final descriptor is
\begin{equation}
g_t=
[\lambda_l\bar{g}_t^{\mathrm{lidar}};
\lambda_n\bar{g}_t^{\mathrm{nav}};
\lambda_e\bar{g}_t^{\mathrm{ego}}],
\quad
\lambda_l+\lambda_n+\lambda_e=1.
\label{eq:risk_fusion}
\end{equation}
Unless otherwise stated, we use equal weights $\lambda_l=\lambda_n=\lambda_e=1/3$.
The descriptor is not used as a reward term; DLIR encodes it to generate the latent-alignment weights in Eq.~\eqref{eq:risk_weight}.

Each observation group and the risk descriptor are encoded by lightweight encoders:
\begin{equation}
\begin{aligned}
e_t^m &= E_m(x_t^m), \quad m\in\mathcal{M},\\
e_t^{\mathrm{risk}} &= E_{\mathrm{risk}}(g_t).
\end{aligned}
\label{eq:group_encoder}
\end{equation}
The encoders are kept separate to preserve the different semantics and scales of kinematic, range, navigation, and risk-derived signals.
% 中文翻译：
% 风险引导的分组编码。
% 将驾驶观测划分为若干语义组 $o_t=\{x_t^m\}_{m\in\mathcal{M}}$，
% 每一组对应一种不同类型的驾驶信息。
% 风险描述符 $g_t$ 由 $\Gamma(o_t)$ 确定性计算，汇总距离接近性和路线一致性线索，不引入额外监督预测目标。
% 在 DLIR 中，$g_t$ 只作为无解码器潜在对齐的组件级引导。
% 风险描述符构造。
% 我们将描述符分解为 LiDAR、导航和自车状态三个组件，分别对应环境接近风险、路线约束风险和车辆状态风险。
% 对于包含 beam 距离 $\{d_{t,i}\}_{i=1}^{N_l}$ 的类 LiDAR 距离感知，接近风险定义为 $\{\exp(-d_{t,i}/\sigma_l)\}_{i=1}^{N_l}$，
% 其中 $\sigma_l$ 控制距离衰减；该形式为近距离障碍物分配更大响应，并抑制远距离回波。
% 导航风险由横向路线误差、航向误差和局部路线曲率组成。
% 当导航接口不直接提供曲率时，曲率由连续导航 waypoint 的航向变化近似为 $\kappa_t\approx |\Delta\psi_t^{\mathrm{nav}}|/\Delta s$。
% 自车状态风险由超出参考速度的速度风险、转向命令接近边界风险和纵向控制接近边界风险组成。
% 编码前，每个风险块按最大-最小范围归一化，并以 $\lambda_l$、$\lambda_n$ 和 $\lambda_e$ 融合为最终描述符。
% 除非另有说明，三个权重取等值 $1/3$。
% 该描述符不作为奖励项使用，而是由 DLIR 编码并用于生成公式~\eqref{eq:risk_weight} 中的潜在对齐权重。
% 每个观测组和风险描述符分别由轻量编码器映射为 $e_t^m$ 和 $e_t^{\mathrm{risk}}$。
% 分开编码用于保留运动学、距离、导航和风险派生信号的不同语义与尺度。

\begin{figure}[!t]
\centering
\includegraphics[width=0.92\linewidth]{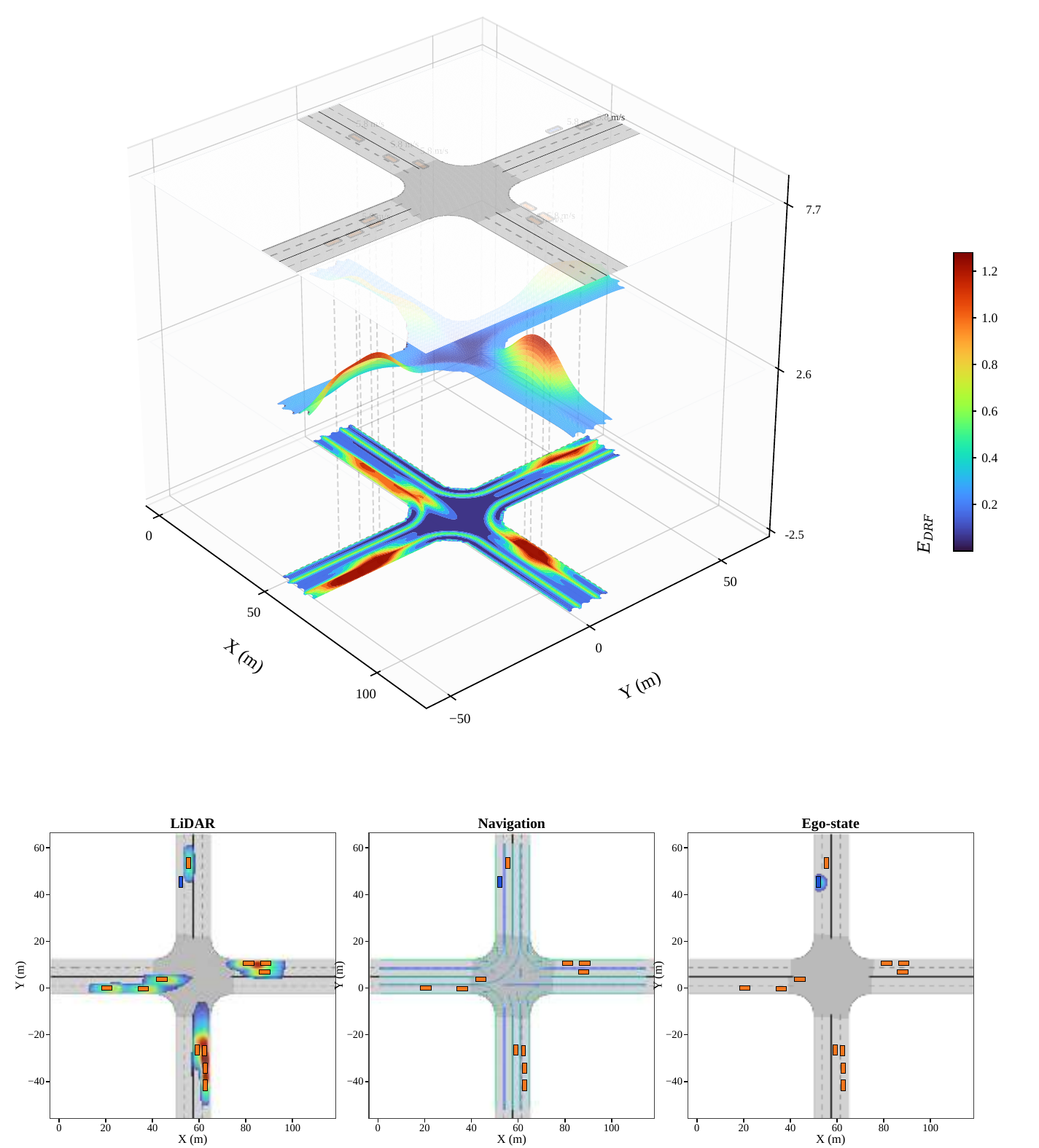}
\caption{DLIR risk descriptor.
The deterministic descriptor combines LiDAR proximity, navigation constraint, and ego-state control-boundary cues, and is used only inside the DLIR representation objective.}
\label{fig:driving_risk_field}
\vspace{-15pt}
\end{figure}
% 图片注释中文翻译：
% DLIR 风险描述符。
% 该确定性描述符结合 LiDAR 接近性、导航约束和自车控制边界线索，并且只在 DLIR 表征目标内部使用。

\noindent\textbf{Latent interaction fusion.}
DLIR obtains an interaction-aware observation embedding by modeling pairwise dependencies among grouped observation embeddings.
For each pair $(m,n)\in\mathcal{P}$, the interaction feature is computed as
\begin{equation}
I_t^{m,n}=
\psi_{m,n}
\left(
e_t^m \odot W_{m,n}e_t^n
\right),
\label{eq:pairwise_interaction}
\end{equation}
where $\odot$ denotes element-wise multiplication and $W_{m,n}$ is a learnable projection.
A context-dependent scalar gate then modulates each pairwise interaction,
\begin{align}
\bar{e}_t &= [e_t^{1};\cdots;e_t^{|\mathcal{M}|};e_t^{\mathrm{risk}}], \label{eq:concat_groups}\\
\alpha_t^{m,n}
&=
\sigma
\left(
w_{m,n}^{\top}\bar{e}_t+b_{m,n}
\right).
\label{eq:interaction_gate}
\end{align}

The final DLIR embedding fuses grouped embeddings, the risk embedding, and gated interaction features:
\begin{equation}
\begin{aligned}
e_t^{\mathrm{DLIR}}
&=
\psi_{\mathrm{fuse}}\!\left(
\left[
\bar{e}_t;
\{\alpha_t^{m,n}I_t^{m,n}\}_{(m,n)\in\mathcal{P}}
\right]\right).
\end{aligned}
\label{eq:dlir_fusion}
\end{equation}
The RSSM posterior is then inferred from the fused embedding:
\begin{equation}
\begin{aligned}
z_t &\sim q_\phi(z_t\mid h_t,e_t^{\mathrm{DLIR}}),\\
s_t^q&=(h_t,z_t).
\end{aligned}
\label{eq:dlir_posterior}
\end{equation}
% 中文翻译：
% DLIR 通过建模不同分组嵌入之间的成对依赖关系获得交互感知观测嵌入。
% 对每一对观测组 $(m,n)$，先通过逐元素乘法和可学习投影得到交互特征 $I_t^{m,n}$。
% 随后，根据包含风险嵌入的整体上下文计算标量门控 $\alpha_t^{m,n}$，用于调节不同交互项的重要性。
% 最终，DLIR 将分组嵌入、风险嵌入和门控交互特征融合为 $e_t^{\mathrm{DLIR}}$，以在 RSSM 后验推断前表达跨组关系。
% 随后，RSSM 根据 $h_t$ 和 $e_t^{\mathrm{DLIR}}$ 推断后验随机状态 $z_t$，得到后验潜在状态 $s_t^q$。

\noindent\textbf{Risk-weighted latent alignment.}
DLIR treats the RSSM posterior state and the interaction-aware observation embedding as two views of the same driving state and projects them into a common space:
\begin{equation}
\begin{aligned}
y_t &= P_s(s_t^q),\\
v_t &= P_e(e_t^{\mathrm{DLIR}}),
\end{aligned}
\label{eq:dlir_projection}
\end{equation}
where $P_s$ and $P_e$ are lightweight projection heads.
For a batch of samples, the normalized cross-correlation matrix between ${y_t}$ and ${v_t}$ is computed as in Eq.~\eqref{eq:bt_corr}.
The risk embedding produces dimension-wise alignment weights:
\begin{equation}
\omega=\mathrm{Norm}\!\left(
\frac{1}{B}\sum_{b=1}^{B}
\mathrm{softplus}(W_\omega e_{t}^{\mathrm{risk},(b)}+b_\omega)
\right),
\label{eq:risk_weight}
\end{equation}
where $\omega_i>0$ and $\frac{1}{D}\sum_i\omega_i=1$.
The DLIR loss is then defined as
\begin{equation}
\mathcal{L}_{\mathrm{DLIR}}
=
\sum_i \omega_i(1-C_{ii})^2
+
\lambda_{\mathrm{off}}
\sum_i\sum_{j\neq i}\sqrt{\omega_i\omega_j}\,C_{ij}^{2}.
\label{eq:dlir_loss}
\end{equation}
The first term performs risk-weighted alignment, and the second reduces redundancy between weighted dimensions.
Applied to $s_t^q$ and $e_t^{\mathrm{DLIR}}$, the loss jointly regularizes grouped encoders, interaction fusion, and RSSM posterior inference.
When all $\omega_i=1$, it reduces to the equal-weight BT-style objective; in DLIR, $\omega$ is risk-conditioned and component-selective.
% 中文翻译：
% 风险加权的潜在对齐。
% DLIR 将 RSSM 后验状态和交互感知观测嵌入视为同一驾驶状态的两个视图，并通过轻量投影头映射到公共表征空间。
% 对一个 batch 内的样本，按照 Barlow Twins 的方式计算 $y_t$ 和 $v_t$ 之间的归一化互相关矩阵。
% 风险嵌入进一步生成维度级对齐权重 $\omega$，其中 $\omega_i>0$ 且平均权重归一化为 1。
% DLIR 损失由两部分组成：
% 第一项执行风险加权对齐，第二项降低加权维度之间的冗余。
% 该损失同时作用于 $s_t^q$ 和 $e_t^{\mathrm{DLIR}}$，联合正则化分组编码器、交互融合模块和 RSSM 后验推断。
% 当所有 $\omega_i=1$ 时，该目标退化为等权重的 BT-style 目标；在 DLIR 中，$\omega$ 由风险描述符条件化，因此具有组件选择性。

\subsection{Residual-Action World Model}
\label{sec:rawm}

RAWM is an action-conditioned RSSM world model with a residual action interface.
It preserves the standard decomposition into posterior inference, prior transition, and prediction heads, but injects control through a joint embedding of the executed command and its residual change.
This converts the temporal correlation in Fig.~\ref{fig:residual_action_horizon} into an explicit representation of local action evolution.
% 中文翻译：
% RAWM 是一个具有残差动作接口的动作条件化 RSSM 世界模型。
% 它保留后验推断、先验转移和预测头组成的标准世界模型分解，但通过执行命令及其残差变化的联合嵌入注入控制信息。
% 这将图~\ref{fig:residual_action_horizon} 中的时间相关性转化为局部动作演化的显式表示。

\noindent\textbf{Model components.}
Let $\mathcal{A}=(-1,1)^d$ be the executed action space.
Given the DLIR posterior state $s_t^q$ in Eq.~\eqref{eq:dlir_posterior}, RAWM uses the following components:
\begin{align}
\textbf{Inv. action:}\quad
u_{t-1}
&= \operatorname{atanh}\!\left(
\operatorname{clip}(a_{t-1},-1+\epsilon,1-\epsilon)\right), \label{eq:rawm_inverse}\\
\textbf{Residual actor:}\quad
\Delta u_t
&\sim \pi_\theta(\Delta u_t\mid s_t^q,u_{t-1}), \label{eq:res_policy}\\
\textbf{Latent update:}\quad
u_t
&= u_{t-1}+\Delta u_t, \label{eq:pre_squash_update}\\
\textbf{Action decode:}\quad
a_t
&= \tanh(u_t), \label{eq:squash}\\
\textbf{Action encode:}\quad
d_t
&= E_{\mathrm{act}}([a_t;\Delta u_t]), \label{eq:joint_act_emb}\\
\textbf{RSSM trans.:}\quad
h_{t+1}
&= f_\phi(h_t,z_t,d_t), \nonumber\\
\textbf{Stoch. prior:}\quad
\hat{z}_{t+1}
&\sim p_\phi(\hat{z}_{t+1}\mid h_{t+1}), \label{eq:rawm_components}\\
\textbf{Prior state:}\quad
s_{t+1}^{p}
&= T_\phi(s_t,d_t), \label{eq:rawm_transition}\\
\textbf{Reward head:}\quad
\hat{\rho}_{t+1}
&=r_\phi(s_{t+1}^{p}), \nonumber\\
\textbf{Continue head:}\quad
\hat{c}_{t+1}
&=C_\phi(s_{t+1}^{p}). \label{eq:rawm_heads}
\end{align}
Here $\epsilon$ is used only for numerical stability near the action boundary.
The induced latent-tanh residual map is
\begin{equation}
\begin{aligned}
F(a_{t-1},\Delta u_t)
&=\tanh\!\left(
\operatorname{atanh}(a_{t-1})+\Delta u_t
\right),\\
&\hspace{1.2em} a_{t-1}\in\mathcal{A}.
\end{aligned}
\label{eq:latent_residual_map}
\end{equation}
% 中文翻译：
% 模型组件。
% 设执行动作空间为 $\mathcal{A}=(-1,1)^d$。
% 给定公式~\eqref{eq:dlir_posterior} 中的 DLIR 后验状态 $s_t^q$，RAWM 由若干紧凑组件组成：
% 先将上一执行动作映射回 latent action space，再由 residual actor 预测 $\Delta u_t$；
% 随后通过 pre-tanh update 和 action decoder 得到执行动作 $a_t$；
% action encoder 将 $a_t$ 与 $\Delta u_t$ 编码为 $d_t$，并输入 RSSM transition；
% 最后，reward head 和 continue head 从预测的 prior state 输出任务相关信号。

Section~\ref{sec:residual_theory} analyzes this map; here the equations define its role in the world model.
% 中文翻译：
% 第~\ref{sec:residual_theory} 节分析该映射的理论性质；此处公式定义其在世界模型中的作用。

\subsection{Structural Analysis of Latent-Tanh Residual Actions}
\label{sec:residual_theory}

We analyze command invariance, interior reachability, and local compactness of the latent-tanh residual map in Eq.~\eqref{eq:latent_residual_map}.
The argument applies elementwise, and vector notation is used for compactness.
Since $\tanh(\cdot)$ produces open-interval actions, the analysis uses $\mathcal{A}=(-1,1)^d$; the clipping in Eq.~\eqref{eq:rawm_inverse} is only a numerical safeguard near the boundary.
Unless otherwise stated, the reachability statement characterizes the parameterization itself and assumes an unrestricted latent residual.

% 中文翻译：
% 下面分析公式~\eqref{eq:latent_residual_map} 中 latent-tanh residual map 的命令不变性、内部可达性和平滑控制假设下的局部紧凑性。
% 该分析逐维成立，向量记号仅用于简洁表达。
% 由于 $\tanh(\cdot)$ 会产生开区间内的动作，理论分析针对 $\mathcal{A}=(-1,1)^d$ 展开；公式~\eqref{eq:rawm_inverse} 中的 clipping 仅用于数值稳定。
% 除非特别说明，可达性结论刻画的是该参数化形式本身，并假设 latent residual 不被人为限制幅值。

\noindent\textbf{Proposition 1 (invariance, reachability, and smoothness).}
For any $a_{t-1}\in\mathcal{A}$, the map in Eq.~\eqref{eq:latent_residual_map} satisfies three properties.
First, a zero residual preserves the previous command:
\begin{equation}
F(a_{t-1},0)=a_{t-1}. 
\label{eq:zero_residual}
\end{equation}
Second, any interior target $a_t^\star\in\mathcal{A}$ is reachable by choosing
\begin{equation}
\Delta u_t^\star
=\operatorname{atanh}(a_t^\star)
-\operatorname{atanh}(a_{t-1}),
\label{eq:target_residual}
\end{equation}
which gives
\begin{equation}
F(a_{t-1},\Delta u_t^\star)=a_t^\star .
\label{eq:rawm_reachability}
\end{equation}
Third, for small $\Delta u_t$,
\begin{align}
a_t-a_{t-1}
&=J(a_{t-1})\Delta u_t
+O(\|\Delta u_t\|^2), \label{eq:rawm_local_continuity}\\
J(a)&=\operatorname{diag}(1-a^2). \nonumber
\end{align}

% 中文翻译：
% 命题 1（不变性、一步可达性和局部平滑性）。
% 对任意上一时刻动作 $a_{t-1}\in\mathcal{A}$，公式~\eqref{eq:latent_residual_map} 中的 latent-tanh residual map 满足三条性质：
% 第一，zero residual 会严格保持上一动作不变；
% 第二，对于任意内部目标动作 $a_t^\star$，都存在一个 latent residual 使其一步到达；
% 第三，小的 latent residual 会通过 tanh 的局部雅可比矩阵诱导平滑的动作变化。

\emph{Proof.}
The invariance property follows directly from the inverse relationship between $\tanh(\cdot)$ and $\operatorname{atanh}(\cdot)$ on $(-1,1)$.
For any target $a_t^\star\in\mathcal{A}$, $\operatorname{atanh}(a_t^\star)$ is finite.
Substituting Eq.~\eqref{eq:target_residual} into the latent-tanh residual map gives Eq.~\eqref{eq:rawm_reachability}.
Finally, applying a first-order Taylor expansion of $\tanh(u_{t-1}+\Delta u_t)$ around $u_{t-1}=\operatorname{atanh}(a_{t-1})$ gives Eq.~\eqref{eq:rawm_local_continuity}.
\hfill$\square$

% 中文翻译：
% 证明。
% 不变性直接来自于 $\tanh(\cdot)$ 和 $\operatorname{atanh}(\cdot)$ 在 $(-1,1)$ 上的互逆关系。
% 对任意目标动作 $a_t^\star\in\mathcal{A}$，$\operatorname{atanh}(a_t^\star)$ 是有限值。
% 将 $\Delta u_t^\star$ 代入 latent-tanh residual map，即可得到一步可达性。
% 最后，在 $u_{t-1}=\operatorname{atanh}(a_{t-1})$ 附近对 $\tanh(u_{t-1}+\Delta u_t)$ 进行一阶 Taylor 展开，即可得到局部平滑性。

Proposition~1 shows that zero residuals preserve the previous command, unrestricted residuals reach any interior target in one step, and small residuals induce smooth local action changes.
If a bounded residual head is used in implementation, the reachability result holds for target actions whose required latent residual lies within the admissible residual range.

% 中文翻译：
% 命题 1 表明，zero residual 可以保持上一控制命令；不受限 residual 可以一步到达任意内部目标动作；小 residual 会诱导平滑局部动作变化。
% 如果实际实现中使用了有界 residual head，则可达性结论适用于所需 latent residual 位于允许范围内的目标动作。

\noindent\textbf{Comparison with a naive action-space residual.}
A naive alternative is to add the residual directly to the executed action,
$a_t=\tanh(a_{t-1}+\Delta u_t)$.
Although this form can cover the interior of $\mathcal{A}$ with unbounded residuals, a zero residual gives
\begin{equation}
\tanh(a_{t-1})\ne a_{t-1}
\quad (a_{t-1}\ne\mathbf{0}).
\label{eq:paper_tanh_drift}
\end{equation}
Repeated zero-residual updates therefore contract each nonzero action component toward zero.
In contrast, the latent-tanh form applies the residual in the pre-tanh action space and preserves the executed command, which is important when maintaining the current steering or throttle is a valid driving decision.

% 中文翻译：
% 与朴素 action-space residual 的比较。
% 一种朴素替代形式是直接将 residual 加到已经执行的动作上，然后再经过 tanh，即
% $a_t=\tanh(a_{t-1}+\Delta u_t)$。
% 尽管该形式在 residual 不受限时也可以覆盖 $\mathcal{A}$ 的内部，但 zero residual 会得到 $\tanh(a_{t-1})$，
% 因而无法保持非零上一动作不变。
% 因此，反复执行 zero-residual update 会使每个非零动作分量逐渐向 0 收缩。
% 相比之下，latent-tanh 形式是在 pre-tanh action space 中施加 residual，因此可以保持执行动作；这对保持当前转向或油门的驾驶决策很重要。

\noindent\textbf{Proposition 2 (local geometric compactness).}
Let $c_\epsilon=\operatorname{atanh}(1-\epsilon)$ and
$\mathcal{U}_\epsilon=[-c_\epsilon,c_\epsilon]^d$ denote the numerically valid latent action domain.
For a finite-domain comparison, an absolute pre-tanh parameterization treats each step as selecting a new latent action $u_t\in\mathcal{U}_\epsilon$.
If a smooth driving segment satisfies $\|u_t-u_{t-1}\|_\infty\le\delta$ with $\delta\ll c_\epsilon$,
then the same local transition is represented by the residual
$\Delta u_t=u_t-u_{t-1}\in[-\delta,\delta]^d$.
The relative hypervolume of this local residual region is
\begin{equation}
\frac{\operatorname{Vol}([-\delta,\delta]^d)}
{\operatorname{Vol}(\mathcal{U}_\epsilon)}
=
\left(\frac{\delta}{c_\epsilon}\right)^d .
\label{eq:rawm_volume_ratio}
\end{equation}

% 中文翻译：
% 命题 2（局部几何紧凑性）。
% 设 $c_\epsilon=\operatorname{atanh}(1-\epsilon)$，并令
% $\mathcal{U}_\epsilon=[-c_\epsilon,c_\epsilon]^d$ 表示数值有效的 latent action domain。
% 为了进行有限域比较，绝对 pre-tanh 动作参数化可以看作每一步都在整个 $\mathcal{U}_\epsilon$ 中重新选择一个 latent action $u_t$。
% 如果某段平滑驾驶过程满足
% $\|u_t-u_{t-1}\|_\infty\le\delta$，且 $\delta\ll c_\epsilon$，
% 则同样的局部动作转移可以由 residual
% $\Delta u_t=u_t-u_{t-1}\in[-\delta,\delta]^d$ 表示。
% 该局部 residual 区域相对于完整 latent action domain 的超体积比为
% $\left(\frac{\delta}{c_\epsilon}\right)^d$。

\emph{Proof.}
Since $\Delta u_t=u_t-u_{t-1}$, the smooth-control condition restricts the residual to $[-\delta,\delta]^d$, whose volume is $(2\delta)^d$.
The reference latent action domain has volume $(2c_\epsilon)^d$.
Taking their ratio gives Eq.~\eqref{eq:rawm_volume_ratio}.
\hfill$\square$

% 中文翻译：
% 证明。
% 由于 $\Delta u_t=u_t-u_{t-1}$，平滑控制条件将 residual 限制在 $[-\delta,\delta]^d$ 内，其体积为 $(2\delta)^d$。
% 参考 latent action domain 的体积为 $(2c_\epsilon)^d$。
% 两者相除即可得到公式~\eqref{eq:rawm_volume_ratio}。

This ratio is a geometric proxy for the local variation represented under the smooth-control assumption, not an action-space restriction, because Proposition~1 preserves one-step interior reachability.

% 中文翻译：
% 该体积比是平滑控制假设下局部动作变化的几何代理，并不意味着动作空间被限制，因为命题 1 保持内部动作的一步可达性。

\noindent\textbf{Corollary 1 (long-horizon residual chains).}
Over a horizon of $K$ steps, residuals accumulate linearly in latent action space:
\begin{equation}
\begin{aligned}
u_{t+K} &= u_t+\sum_{k=1}^{K}\Delta u_{t+k},\\
a_{t+K} &= \tanh(u_{t+K}).
\end{aligned}
\label{eq:rawm_long_horizon}
\end{equation}
If the smoothness condition holds at each step, the residual sequence lies in $[-\delta,\delta]^{Kd}$, whereas an unconstrained absolute-action sequence lies in $\mathcal{U}_\epsilon^K$.
Their relative hypervolume is
\begin{equation}
\frac{\operatorname{Vol}([-\delta,\delta]^{Kd})}
     {\operatorname{Vol}(\mathcal{U}_\epsilon^K)}
=
\left(\frac{\delta}{c_\epsilon}\right)^{Kd}.
\label{eq:rawm_chain_volume}
\end{equation}
Thus, under a smooth-control assumption, coherent maneuvers can be described by compact residual chains while decoded commands remain in the original tanh action space.
This view is consistent with Fig.~\ref{fig:residual_action_horizon}, where residual-action chains occupy 9.1$\times$ fewer cells than absolute-action chains.

% 中文翻译：
% 推论 1（长时域 residual action chains）。
% 在 $K$ 步时域内，residual 会在 latent action space 中线性累积。
% 如果平滑性条件在每一步都成立，则 residual sequence 位于 $[-\delta,\delta]^{Kd}$ 中；
% 而无约束的绝对动作序列则位于 $\mathcal{U}_\epsilon^K$ 中。
% 二者的相对超体积为
% $\left(\frac{\delta}{c_\epsilon}\right)^{Kd}$。
% 因此，在平滑控制假设下，连续驾驶机动可以由紧凑 residual chains 描述，同时解码后的执行命令仍位于原始 tanh action space 内。
% 这一视角与图~\ref{fig:residual_action_horizon} 中的经验统计一致：residual-action chains 相比 absolute-action chains 占据的网格数量减少了 9.1 倍。

\subsection{Residual-Action Chain Contrastive Learning}
\label{sec:arccl}

\begin{figure}[t]
\centering
\includegraphics[width=0.86\linewidth]{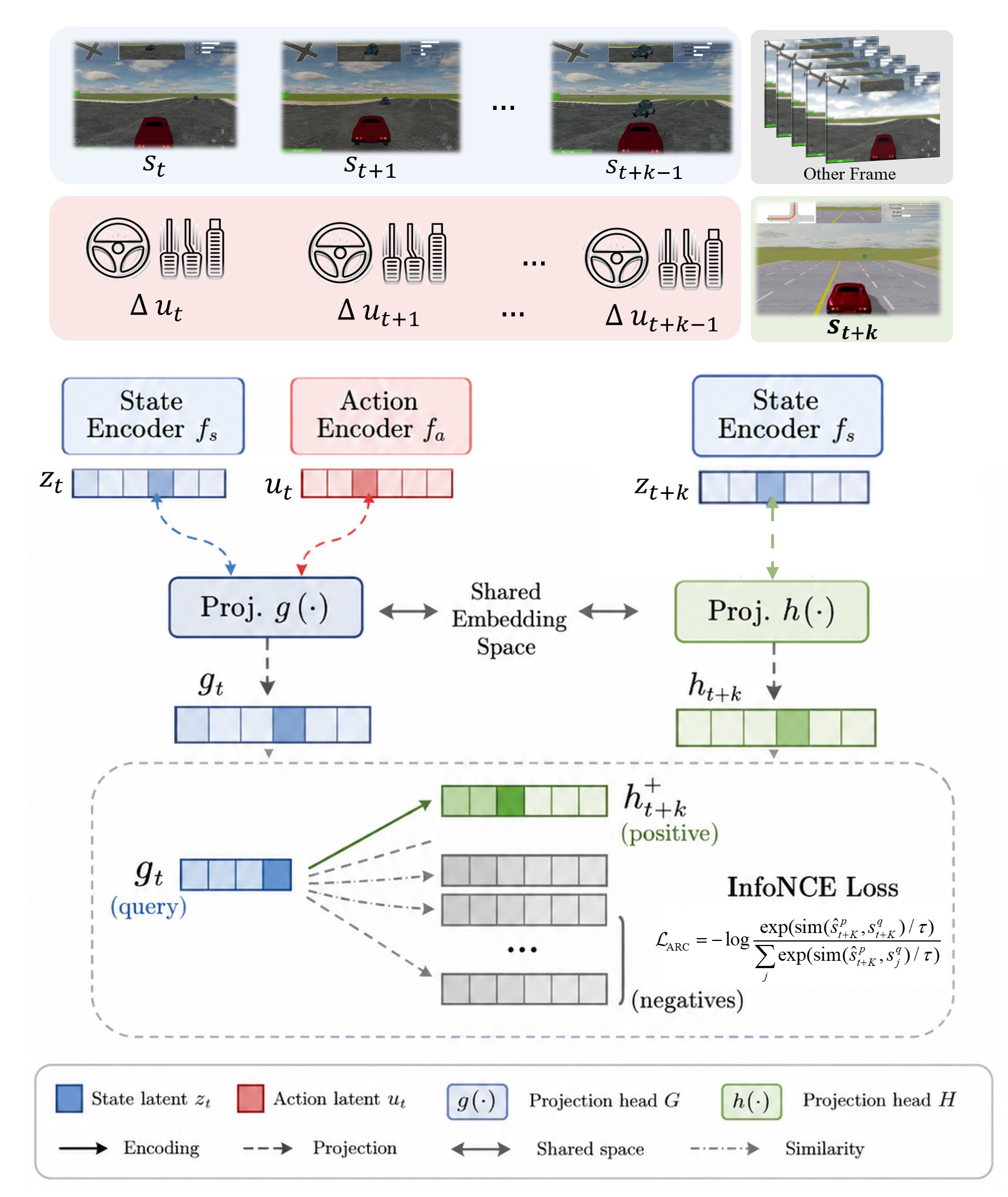}
\caption{Residual-action chain contrast.
Starting from the shared posterior state, the transition model is recursively applied with one joint action embedding at each step, and the final predicted latent state is aligned with the true future posterior by InfoNCE.}
\label{fig:arc_framework}
\vspace{-4pt}
\end{figure}

ARC-CL regularizes the world model with a contrastive objective on multi-step rollouts driven by successive joint action embeddings, as shown in Fig.~\ref{fig:arc_framework}.
% 中文翻译：
% ARC-CL 将对比目标作用于由逐步联合动作嵌入驱动的多步 rollout，以正则化世界模型，如图~\ref{fig:arc_framework} 所示。

For a horizon $K$, ARC-CL uses the ordered list of per-step joint action embeddings
\begin{align}
\mathcal{D}_t^K &= (d_t,d_{t+1},\ldots,d_{t+K-1}), \label{eq:action_embedding_sequence}\\
\hat{s}_{t}^{p} &= s_t^q, \nonumber\\
\hat{s}_{t+k+1}^{p} &= T_\phi(\hat{s}_{t+k}^{p},d_{t+k}), \quad k=0,\ldots,K-1. \label{eq:kstep_recursive}
\end{align}
The predicted future state is $\hat{s}_{t+K}^{p}$.
The list $\mathcal{D}_t^K$ specifies recursive rollout inputs only; it is not compressed into a chain-level feature or fed to the transition as a single input.
The rollout uses prior dynamics and per-step action embeddings without future observations.
For world-model and ARC-CL training, each $d_{t+k}$ is constructed from the replayed action $a_{t+k}$ and residual command $\Delta u_{t+k}$ produced by the behavior actor.
It is not recomputed with the current actor, keeping the action embedding consistent with the realized replay transition.
% 中文翻译：
% 对于长度为 $K$ 的 horizon，ARC-CL 使用逐步联合动作嵌入的有序列表 $\mathcal{D}_t^K$。
% 从当前后验状态 $s_t^q$ 出发，模型每一步输入一个 $d_{t+k}$，递归调用 RAWM 转移模型，得到 $K$ 步后的先验预测状态。
% $\mathcal{D}_t^K$ 仅指定递归 rollout 的逐步输入，不会被压缩为链级特征，也不会作为单个输入送入 transition。
% rollout 只使用先验动力学和逐步动作嵌入，不访问未来观测。
% 在世界模型和 ARC-CL 训练中，$d_{t+k}$ 使用 replay 中存储的执行动作和行为 actor 产生的 residual command，不由当前 actor 重新计算，以保持 off-policy 一致性。
% 图片注释中文翻译：残差动作链对比学习：从共享后验出发，逐步递归rollout，通过InfoNCE对齐预测未来潜状态与真实后验。

The predicted future state $\hat{s}_{t+K}^{p}$ is aligned with the future posterior $s_{t+K}^{q}$ through InfoNCE~\cite{Oord2018CPC}.
For a batch $B$, define
\begin{equation}
\ell_{b,b'}=
\mathrm{sim}\!\left(s_{t+K}^{q,(b)},\hat{s}_{t+K}^{p,(b')}\right).
\label{eq:arc_score}
\end{equation}
Positive pairs use $b'=b$, and negatives use other trajectory segments.
The contrastive loss is
\begin{equation}
\mathcal{L}_{\mathrm{ARC}}
=-\frac{1}{|B|}\sum_{b\in B}
\log
\frac{\exp(\ell_{b,b}/\tau)}
{\sum_{b'\in B}\exp(\ell_{b,b'}/\tau)} .
\label{eq:arc_loss}
\end{equation}
Here $\mathrm{sim}(\cdot,\cdot)$ is cosine similarity and $\tau$ is a temperature.
The loss updates $T_\phi$, $E_{\mathrm{act}}$, and DLIR through $s_t^q$; the residual policy is updated separately by the latent actor objective during imagined rollouts.
% 中文翻译：
% 预测的未来状态 $\hat{s}_{t+K}^{p}$ 通过 InfoNCE 与真实未来后验状态 $s_{t+K}^{q}$ 对齐。
% 在一个 batch 中，先计算真实未来后验状态与各预测未来状态之间的相似度分数 $\ell_{b,b'}$。
% 正样本对应 $b'=b$，负样本来自其他轨迹片段；相似度函数为余弦相似度，$\tau$ 为温度参数。
% 该损失通过 $s_t^q$ 更新转移模型、动作嵌入和 DLIR；residual policy 由 imagined rollout 中的 actor objective 单独更新。

\subsection{Optimization and Policy Learning}
\label{sec:loss}

The world model is trained with task prediction, RSSM regularization, decoder-free representation learning, and multi-step action contrast:
\begin{equation}
\begin{aligned}
\mathcal{L}_{\mathrm{WM}}
=&\; \mathcal{L}_{\mathrm{reward}}
+ \mathcal{L}_{\mathrm{continue}}
+ \beta_{\mathrm{dyn}}\mathcal{L}_{\mathrm{dyn}}
+ \beta_{\mathrm{rep}}\mathcal{L}_{\mathrm{rep}}\\
&+ \lambda_{\mathrm{DLIR}}\mathcal{L}_{\mathrm{DLIR}}
+ \lambda_{\mathrm{ARC}}\mathcal{L}_{\mathrm{ARC}} .
\end{aligned}
\label{eq:total_loss}
\end{equation}
The reward and continuation heads are trained by negative log likelihood, while the two RSSM regularizers are
\begin{align}
\mathcal{L}_{\mathrm{dyn}}
&=D_{\mathrm{KL}}\!\left(q_\phi(z_t\mid h_t,e_t^{\mathrm{DLIR}})
\Vert p_\phi(\hat{z}_t\mid h_t)\right), \label{eq:loss_dyn}\\
\mathcal{L}_{\mathrm{rep}}
&=D_{\mathrm{KL}}\!\left(q_\phi(z_t\mid h_t,e_t^{\mathrm{DLIR}})
\Vert \mathcal{N}(0,I)\right). \label{eq:loss_rep}
\end{align}
The proposed losses act on the same posterior and prior states as the RSSM terms, enforcing compactness and multi-step predictive consistency in one latent space.
% 中文翻译：
% 世界模型由任务预测、RSSM 正则化、无解码器表征学习和多步动作对比共同训练。
% 总损失包含奖励预测、继续信号预测、动力学 KL、表征 KL、DLIR 冗余降低损失和 ARC-CL 对比损失。
% 奖励和继续信号预测头使用负对数似然训练，RSSM 正则项约束后验与先验以及后验与标准先验分布之间的关系。
% 所提出的 $\mathcal{L}_{DLIR}$ 和 $\mathcal{L}_{ARC}$ 作用于同一组后验和先验状态，使紧凑性和多步预测一致性在同一 latent space 中被优化。

\noindent\textbf{Latent actor-critic learning.}
\label{sec:ac}
Following Dreamer-style latent imagination~\cite{Hafner2020Dreamer,Hafner2023DreamerV3}, the actor and critic are optimized on trajectories generated by the learned world model.
Imagined rollouts start from replay posterior states and use the residual action recursion in Eqs.~\eqref{eq:res_policy}, \eqref{eq:pre_squash_update}, and~\eqref{eq:squash}.
The critic and actor objectives are
\begin{align}
\mathcal{L}_{\mathrm{critic}}(\psi)
&=\frac{1}{2}\mathbb{E}_{\tau_{\mathrm{imag}}}
\left[(v_\psi(s_t)-V_t^\lambda)^2\right], \label{eq:critic_loss}\\
\mathcal{L}_{\mathrm{actor}}(\theta)
&=-\mathbb{E}_{\tau_{\mathrm{imag}}}
\left[\sum_{k=0}^{H_{\mathrm{imag}}-1}\gamma^k G_{t+k}\right], \label{eq:actor_loss}
\end{align}
where $G_{t+k}=v_\psi(\hat{s}_{t+k})+\eta\,\mathcal{H}(\pi_\theta(\cdot\mid\hat{s}_{t+k},u_{t+k-1}))$.
During imagination, $u_{t+k}=u_{t+k-1}+\Delta u_{t+k}$ is updated together with the latent state.
% 中文翻译：
% 遵循 Dreamer 式潜在想象，actor 和 critic 在学习到的世界模型生成的轨迹上优化。
% 想象 rollout 从 replay 后验状态开始，并使用 RAWM 的残差动作递推。
% critic 最小化 TD($\lambda$) 目标误差，actor 最大化带熵正则的想象回报。
% 在想象过程中，pre-tanh 动作 $u_{t+k}$ 与 latent state 一起递推更新，从而保持残差动作参数化的一致性。

\noindent\textbf{Training procedure.}
\label{sec:algorithm}
Algorithm~\ref{alg:lidarad} summarizes data collection, world-model learning, and latent actor-critic optimization.
The LIDAR-AD replay buffer stores the shared environment tuple $(o_t,a_t,\rho_t,c_t)$ and additionally records the actor-produced residual $\Delta u_t$ required by RAWM and ARC-CL.
During world-model and ARC-CL updates, $\Delta u_t$ is read from replay rather than recomputed by the current actor.
Non-residual baselines store only the standard transition tuple under the same observation interface, action space, reward function, data budget, and evaluation protocol.
% 中文翻译：
% 算法~\ref{alg:lidarad} 总结数据收集、世界模型学习和潜在 actor-critic 优化。
% LIDAR-AD replay buffer 存储共有环境交互元组 $(o_t,a_t,\rho_t,c_t)$，并额外记录 RAWM 和 ARC-CL 所需的 actor residual $\Delta u_t$。
% 世界模型和 ARC-CL 更新时，$\Delta u_t$ 从 replay 中读取，而不是由当前 actor 重新计算。
% 非 residual baseline 只存储标准 transition tuple，并保持相同观测接口、动作空间、奖励、数据预算和评估协议。

\begin{algorithm}[t]
  \caption{LIDAR-AD Training Procedure}
  \label{alg:lidarad}
  \begin{algorithmic}[1]
    \REQUIRE Environment, ARC-CL rollout horizon $K$, imagination horizon $H_{\mathrm{imag}}$, training steps $N$.
    \STATE Initialize world model $\phi$, DLIR encoders, action encoder $E_{\mathrm{act}}$, actor $\theta$, critic $\psi$, and replay buffer $\mathcal{D}$.
    \STATE Set $a_{-1}\leftarrow \mathbf{0}$ and $s_0\leftarrow \mathbf{0}$.
    \FOR{$n=1$ to $N$}
      \STATE Observe $o_t=(x_t^{\mathrm{ego}},x_t^{\mathrm{lidar}},x_t^{\mathrm{nav}})$.
      \STATE Compute $e_t^{\mathrm{DLIR}}$ with DLIR and infer $s_t^q$ by Eq.~\eqref{eq:dlir_posterior}.
      \STATE Compute $u_{t-1}$ by Eq.~\eqref{eq:rawm_inverse}; sample $\Delta u_t\sim\pi_\theta(\cdot\mid s_t^q,u_{t-1})$.
      \STATE Compute $u_t=u_{t-1}+\Delta u_t$ and $a_t=\tanh(u_t)$; execute $a_t$ and receive reward $\rho_t$.
      \STATE Store $(o_t,a_t,\Delta u_t,\rho_t,c_t)$ in $\mathcal{D}$.
      \IF{$\mathcal{D}$ has enough sequences}
        \STATE Sample a trajectory batch, including stored $(a_t,\Delta u_t)$, and infer posterior states $\{s_t^q\}$.
        \STATE Construct $d_t=E_{\mathrm{act}}([a_t;\Delta u_t])$ from replayed actions and residuals; update RSSM priors with $T_\phi(s_t,d_t)$.
        \STATE Run $K$-step prior rollouts using the ordered per-step embeddings $\mathcal{D}_t^K$; compute $\mathcal{L}_{\mathrm{DLIR}}$, $\mathcal{L}_{\mathrm{ARC}}$, and $\mathcal{L}_{\mathrm{WM}}$.
        \STATE Update world-model parameters by minimizing $\mathcal{L}_{\mathrm{WM}}$.
        \STATE Roll out imagined trajectories from replay posterior states using the residual actor.
        \STATE Update critic with Eq.~\eqref{eq:critic_loss} and actor with Eq.~\eqref{eq:actor_loss}.
      \ENDIF
    \ENDFOR
    \RETURN Trained world model, actor, and critic.
  \end{algorithmic}
\end{algorithm}

\section{Experimental Evaluation}
\label{sec:exp_setup}

% 写作指导：
% 本节按照 IEEE 风格组织为两个主要部分：A. Experiment Setup 和 B. Experimental Results。
% A 部分只说明实验环境、基线与指标、实现细节和实验设置；B 部分由 sec06_exp_results.tex 给出。
% 不在设置部分展开消融算法细节，也不保留没有结果支撑的指标或占位超参数表。

\subsection{Experiment Setup}
\label{sec:experiment_setup}

\subsubsection{Simulation Environment}
\label{sec:setup_metadrive}
\label{sec:setup_spaces}

\begin{figure}[t]
  \centering
  \includegraphics[width=\linewidth]{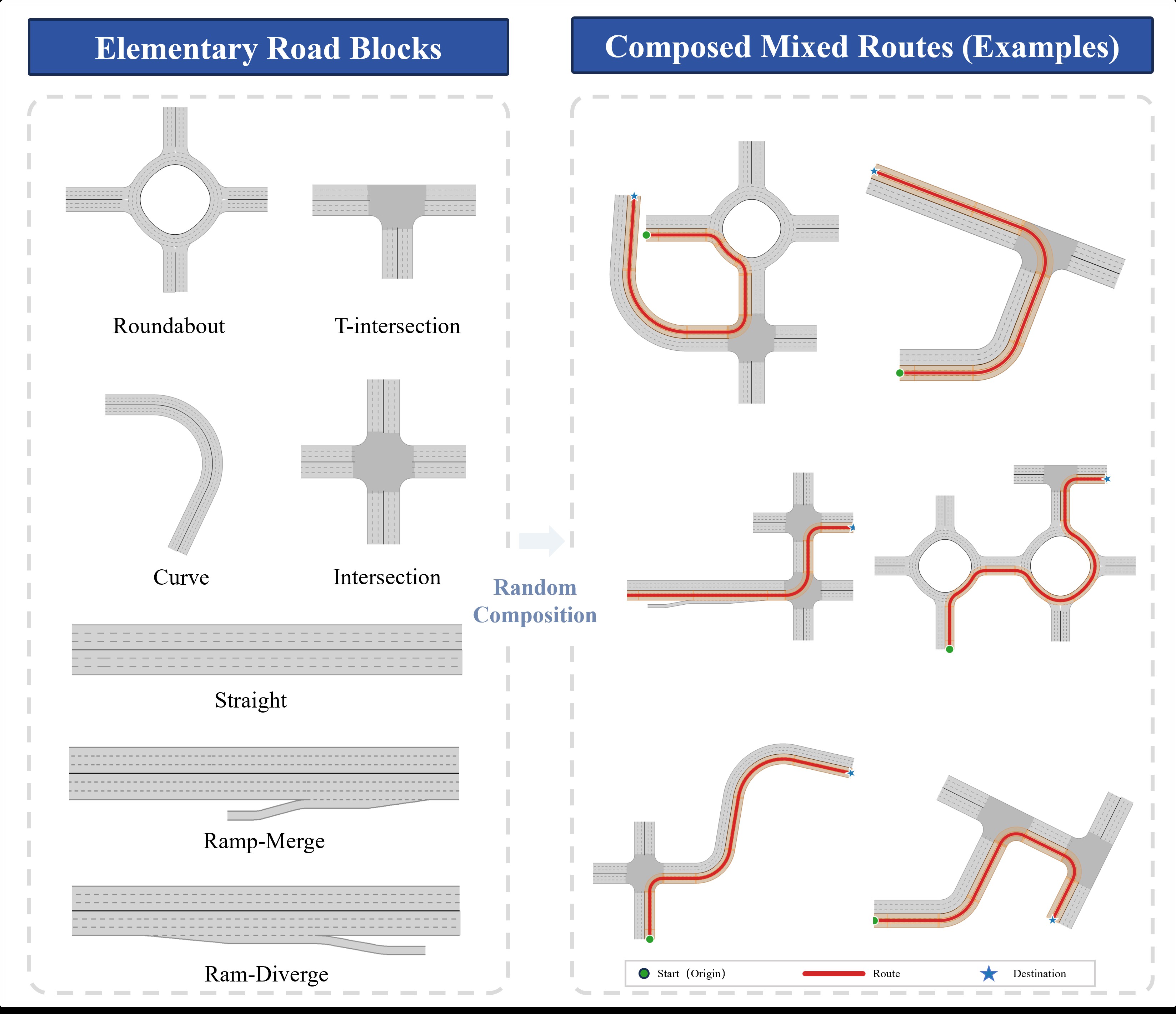}
  \caption{Scenario generation in MetaDrive.
  Elementary road blocks are randomly composed into mixed driving routes, enabling evaluation across diverse intersections, roundabouts, curves, and multi-road layouts under a unified closed-loop simulator interface.}
  \label{fig:environment_examples}
  \vspace{-4pt}
\end{figure}

We use MetaDrive~\cite{Li2022MetaDrive} as the unified closed-loop simulation interface for both procedurally generated scenarios and real-world log-based scenarios.
MetaDrive provides controllable road generation, traffic density, vehicle dynamics, and LiDAR-like range observations, enabling controlled evaluation under diverse geometry and interaction patterns.
To evaluate transfer under real-world traffic layouts, nuPlan logs~\cite{Caesar2021NuPlan} are converted into MetaDrive-compatible scenarios through the official ScenarioNet middleware~\cite{Li2023ScenarioNet}.
This setting allows policies trained and evaluated in a unified simulator interface to be tested on log-reconstructed urban scenarios, but it is not the official nuPlan closed-loop protocol and does not report official nuPlan planning metrics.
Fig.~\ref{fig:environment_examples} shows representative road blocks and composed routes used for scenario randomization.
% 中文翻译：
% 我们使用 MetaDrive 作为统一的闭环仿真接口，同时评估程序化生成场景和真实日志驱动场景。
% MetaDrive 提供可控道路生成、交通密度、车辆动力学和类 LiDAR 距离观测，适合在不同道路几何和交互模式下进行可控评估。
% 为评估真实交通布局下的迁移性，nuPlan 日志通过官方 ScenarioNet 中间件转换为 MetaDrive 可读取的场景。
% 因此，策略可以在统一仿真接口下训练和评估，同时接受日志重建城市场景测试；但这不是官方 nuPlan 闭环协议，也不报告官方 nuPlan planning metrics。
% 图~\ref{fig:environment_examples} 展示了用于场景随机化的代表性道路块和组合路线。

% 图片注释中文翻译：
% MetaDrive 中的场景生成。
% 基础道路块被随机组合为混合驾驶路线，从而在统一闭环仿真接口下评估交叉口、环岛、弯道和多路段布局等多样化场景。

Each observation consists of ego-vehicle state, LiDAR-like range sensing, and navigation information.
The same observation interface is used for all learning-based methods to ensure a controlled comparison across algorithms and datasets.
The action is a two-dimensional continuous command,
\begin{equation}
  a_t = [a_{\mathrm{acc}}, a_{\mathrm{steer}}]^\top \in [-1,1]^2,
  \label{eq:setup_act}
\end{equation}
where $a_{\mathrm{acc}}$ denotes normalized longitudinal control and $a_{\mathrm{steer}}$ denotes normalized steering control.
For LIDAR-AD, this executed action is generated by the latent-tanh residual action parameterization in Section~\ref{sec:rawm}.
% 中文翻译：
% 每个观测由自车状态、类 LiDAR 距离感知和导航信息组成。
% 所有学习式方法使用相同的观测接口，以保证不同算法和数据集之间的可控比较。
% 动作为二维连续控制命令，其中 $a_{\mathrm{acc}}$ 表示归一化纵向控制，$a_{\mathrm{steer}}$ 表示归一化转向控制。
% 对 LIDAR-AD 而言，该执行动作由第~\ref{sec:rawm} 节中的 latent-tanh residual action 参数化生成。

\noindent\textbf{Reward design.}
The closed-loop reward follows the standard driving objective used in the simulator and is kept identical for all methods.
At each step, the scalar reward is composed of route progress, speed incentive, goal completion, and safety penalties:
\begin{equation}
\rho_t =
  w_p\Delta l_t + w_v\bar{v}_t + w_g\mathbb{I}_{\mathrm{goal}}
  - w_c\mathbb{I}_{\mathrm{col}} - w_o\mathbb{I}_{\mathrm{off}},
\label{eq:exp_reward}
\end{equation}
where $\Delta l_t$ is route progress, $\bar{v}_t$ is normalized speed, and $\mathbb{I}_{\mathrm{goal}}$, $\mathbb{I}_{\mathrm{col}}$, and $\mathbb{I}_{\mathrm{off}}$ indicate goal arrival, collision, and off-road events, respectively.
The coefficients are fixed across all methods, so the reported episode reward is directly comparable.
% 中文翻译：
% 奖励设计。闭环奖励遵循仿真器中的标准驾驶目标，并对所有方法保持一致。
% 每一步奖励由路线前进、速度激励、到达目标奖励和安全惩罚组成。
% 其中 $\Delta l_t$ 表示路线前进量，$\bar{v}_t$ 表示归一化速度，三个指示函数分别表示到达目标、碰撞和偏离道路事件。
% 所有方法使用相同奖励系数，因此报告的 episode reward 可以直接比较。
%
% 段落大意：
% 本小节说明统一仿真接口、真实数据导入、观测动作空间和奖励函数设计。

\subsubsection{Baselines and Evaluation Metrics}
\label{sec:setup_baselines}
\label{sec:setup_metrics}

\noindent\textbf{Baselines.}
We compare LIDAR-AD with the following rule-based, model-free, and model-based methods:
\begin{itemize}[leftmargin=*, itemsep=0.3ex]
  \item \textbf{IDM+MOBIL}: a non-learning control baseline that combines the Intelligent Driver Model~\cite{Treiber2000IDM} with the MOBIL lane-change rule~\cite{Kesting2007MOBIL}.
  \item \textbf{SAC}~\cite{Haarnoja2018SAC}: a model-free off-policy RL algorithm for continuous control.
  \item \textbf{PPO}~\cite{Schulman2017PPO}: a model-free on-policy RL algorithm with a clipped policy update.
  \item \textbf{DreamerV3}~\cite{Hafner2023DreamerV3}: the standard RSSM-based latent world-model baseline.
  \item \textbf{R2-Dreamer}~\cite{Morihira2026R2Dreamer}: a decoder-free world-model baseline and the closest reference to our representation-learning setting.
\end{itemize}
All learning-based baselines use the same observation interface, action space, reward, and evaluation budget as LIDAR-AD.
Module ablations are reported separately in Section~\ref{sec:exp_results} to isolate the contributions of DLIR, ResAct, RAWM, and ARC-CL.
% 中文翻译：
% 基线。我们将 LIDAR-AD 与以下规则式、免模型和基于模型的方法进行比较：
% IDM+MOBIL 是结合 IDM 跟车模型和 MOBIL 换道规则的非学习控制基线；
% SAC 是免模型离策略连续控制算法；PPO 是免模型在线策略算法；
% DreamerV3 是标准 RSSM 潜在世界模型基线；
% R2-Dreamer 是无解码器世界模型基线，也是与本文表征学习设定最接近的参考方法。
% 所有学习式基线与 LIDAR-AD 使用相同观测接口、动作空间、奖励和评估预算。
% 模块消融在第~\ref{sec:exp_results} 节单独报告。

\noindent\textbf{Evaluation metrics.}
We report only metrics used in the quantitative results.
For compact table presentation, each metric is first introduced by its full name followed by the abbreviation used in the tables:
\begin{itemize}[leftmargin=*, itemsep=0.3ex]
  \item \textbf{Reward (RW)}: the undiscounted episode return under Eq.~\eqref{eq:exp_reward}.
  \item \textbf{Success rate (SR)}: the percentage of episodes that reach the navigation goal without collision or off-road termination.
  \item \textbf{Route completion (RC)}: the percentage of the planned route completed before termination.
  \item \textbf{Average speed (AS)}: the mean driving speed over an episode, used to measure driving efficiency.
  \item \textbf{Steering delta (SD)} and \textbf{longitudinal-control delta (LCD)}: the mean absolute change between consecutive steering or longitudinal commands, used to measure control smoothness.
  \item \textbf{Normalized AUC at 1M steps (nAUC@1M)}: the normalized area under the training curve over the first one million environment steps, used for sample-efficiency comparison in ablations.
\end{itemize}
% 中文翻译：
% 评估指标。我们只报告定量结果中实际使用的指标：
% 表格中统一使用缩写：RW 为公式~\eqref{eq:exp_reward} 下的无折扣 episode return；
% SR 表示无碰撞、无偏离道路到达导航目标的 episode 百分比；
% RC 表示终止前完成的规划路线比例；
% AS 表示平均速度，用于衡量驾驶效率；
% SD 和 LCD 分别为连续转向或纵向控制命令之间的平均绝对变化，用于衡量控制平滑性；
% nAUC@1M 为前一百万环境步训练曲线的归一化面积，用于消融实验中的样本效率比较。
%
% 段落大意：
% 本小节参考 IEEE 实验设置常见写法，将 baselines 和 metrics 分成两个加粗段落，并用分点定义各方法和指标。

\subsubsection{Implementation Details}
\label{sec:setup_impl}

\begin{figure}[t]
  \centering
  \includegraphics[width=0.95\linewidth]{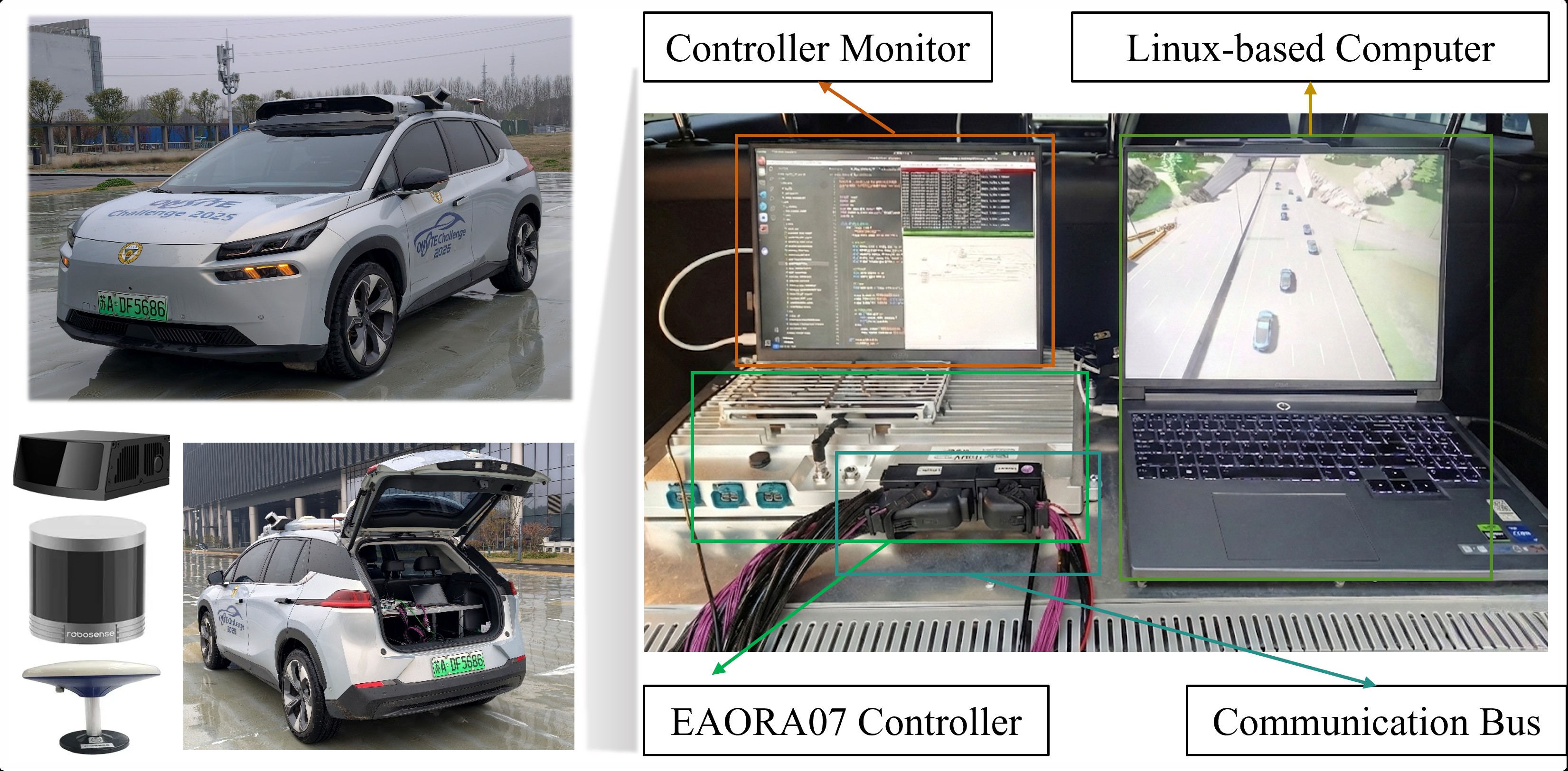}
  \caption{Vehicle platform for real-time feasibility testing.
  The trained policy is deployed on an instrumented vehicle platform to examine inference-time execution on physical hardware, while quantitative closed-loop evaluation is conducted in MetaDrive and nuPlan-derived scenarios.}
  \label{fig:real_time_platform}
  \vspace{-4pt}
\end{figure}

LIDAR-AD is implemented in PyTorch with an RSSM backbone following the DreamerV3 training paradigm.
Compared with DreamerV3, LIDAR-AD replaces the reconstruction-based observation encoder with DLIR, conditions the RSSM transition on the RAWM joint action embedding $d_t$, and adds ARC-CL to the world-model objective.
The DLIR risk descriptor $g_t$ is deterministically computed from structured driving observations as in Eqs.~\eqref{eq:risk_descriptor_blocks}--\eqref{eq:risk_fusion}.
The residual actor predicts $\Delta u_t$ in the pre-tanh action space, and the action embedding network receives $[a_t;\Delta u_t]$ at each transition step.
The ARC-CL rollout horizon is set to $K=10$ unless otherwise stated.
All methods are trained with Adam under matched compute and data budgets.
For real-time feasibility checking, the trained policy is also deployed on an instrumented vehicle platform for inference-time testing, as shown in Fig.~\ref{fig:real_time_platform}.
This deployment is used only to examine execution feasibility on physical hardware; all closed-loop driving performance metrics reported below are obtained from the simulator and nuPlan-derived scenarios.
% 中文翻译：
% LIDAR-AD 使用 PyTorch 实现，RSSM 主干遵循 DreamerV3 训练范式。
% 与 DreamerV3 相比，LIDAR-AD 用 DLIR 替代基于重建的观测编码器，使用 RAWM 的联合动作嵌入 $d_t$ 条件化 RSSM 转移，并在世界模型目标中加入 ARC-CL。
% DLIR 风险描述符 $g_t$ 按公式~\eqref{eq:risk_descriptor_blocks}--\eqref{eq:risk_fusion} 从结构化驾驶观测中确定性计算。
% 残差 actor 在 tanh 之前动作空间中预测 $\Delta u_t$，动作嵌入网络在每个转移步接收 $[a_t;\Delta u_t]$。
% 除非特别说明，ARC-CL 的 rollout horizon 设为 $K=10$。
% 所有方法在匹配的计算和数据预算下使用 Adam 训练。
% 为检查实时执行可行性，训练后的策略还部署到实车平台上进行推理时间测试，如图~\ref{fig:real_time_platform} 所示。
% 该部署仅用于检查物理硬件上的执行可行性；下文报告的闭环驾驶性能均来自仿真器和 nuPlan 派生场景。

% 图片注释中文翻译：
% 用于实时性可行性测试的实车平台。
% 训练后的策略被部署到带传感器和计算设备的车辆平台上，以检查物理硬件上的推理执行；
% 定量闭环评估仍在 MetaDrive 和 nuPlan 派生场景中完成。

\subsubsection{Experiment Settings}
\label{sec:setup_scenarios}

Training is conducted on randomized MetaDrive maps with varying road layouts and traffic configurations at the nominal traffic density of 0.20.
Evaluation is performed on four representative scenario sets used in the main results: mixed scene, roundabout, T-intersection, and nuPlan-derived log-reconstructed scenarios imported through ScenarioNet.
To evaluate traffic-density generalization from this nominal setting, we additionally test mixed, T-intersection, and roundabout scenarios at densities of 0.10, 0.20, and 0.30.
All learned methods are trained and evaluated under the same data-collection budget and randomization protocol.
Unless otherwise stated, results are reported as mean $\pm$ standard deviation across $N_{\mathrm{run}}$ independent training runs; each trained policy is evaluated on the same fixed set of evaluation episodes for the corresponding scenario.
% 中文翻译：
% 训练在随机 MetaDrive 地图上进行，包含变化的道路布局和交通配置，训练交通密度为 0.20。
% 主结果在四个代表性场景集合上报告：mixed scene、roundabout、T-intersection 和通过 ScenarioNet 导入的 nuPlan 派生日志重建场景。
% 为评估相对训练密度的泛化，我们还在 0.10、0.20 和 0.30 交通密度下测试 mixed、T-intersection 和 roundabout 场景。
% 所有学习式方法使用相同的数据收集预算和随机化协议。
% 除非特别说明，结果以 $N_{\mathrm{run}}$ 个独立训练 runs 的均值 $\pm$ 标准差报告；每个训练后的策略在对应场景的固定 evaluation episodes 集合上评估。
% TODO: Replace $N_{\mathrm{run}}$ with the actual number of independent training runs before submission.
%
% 段落大意：
% 本小节说明训练场景、主评估基准和分布偏移测试设置。

\subsection{Experimental Results}
\label{sec:exp_results}

% 写作指导：
% 本节按证据层级组织：整体性能 -> 学习效率 -> 模块贡献 -> 时域敏感性 -> 分布外密度泛化。
% 只报告定量结果和直接支撑结论的图表；机制解释和定性效果图放在 Discussion。

\subsubsection{Overall Closed-Loop Performance}

Table~\ref{tab:overall_performance} reports closed-loop performance on mixed scene, roundabout, T-intersection, and nuPlan-derived log-reconstructed scenarios.
LIDAR-AD achieves the highest reward in all scenarios and the highest success rate among learning-based methods.
Relative to R2-Dreamer, it improves success rate by 4.42, 2.68, 4.18, and 6.33 percentage points on mixed scene, roundabout, T-intersection, and nuPlan-derived scenarios, respectively, increasing the average success rate from 90.40\% to 94.81\%.
Although IDM obtains a slightly higher success rate on nuPlan-derived scenarios, its substantially lower reward indicates more conservative behavior.
These results show that LIDAR-AD improves task completion while preserving closed-loop driving efficiency.
% 中文翻译：
% 表~\ref{tab:overall_performance} 报告 mixed scene、roundabout、T-intersection 和 nuPlan 派生日志重建场景上的闭环驾驶性能。
% LIDAR-AD 在所有场景中均取得最高 reward，并在学习式方法中取得最高 success rate。
% 相比 R2-Dreamer，LIDAR-AD 在四个场景中的成功率分别提升 4.42、2.68、4.18 和 6.33 个百分点，平均成功率从 90.40\% 提升到 94.81\%。
% 尽管 IDM 在 nuPlan 派生场景上成功率略高，但其 reward 明显较低，说明其行为更保守。
% 这些结果表明 LIDAR-AD 在提升任务完成率的同时保持了闭环驾驶效率。

\begin{table*}[!t]
  \centering
  \caption{Overall Driving Performance Across MetaDrive and nuPlan-Derived Scenarios}
  \label{tab:overall_performance}
  \setlength{\tabcolsep}{7.2pt}
  \footnotesize
  \resizebox{\textwidth}{!}{%
  \begin{tabular}{lcccccccc}
    \toprule
    \multirow{2}{*}{\textbf{Method}} &
    \multicolumn{2}{c}{\textbf{Mixed Scene}} &
    \multicolumn{2}{c}{\textbf{Roundabout}} &
    \multicolumn{2}{c}{\textbf{T-Intersection}} &
    \multicolumn{2}{c}{\textbf{nuPlan-Derived}} \\
    \cmidrule(lr){2-3}\cmidrule(lr){4-5}\cmidrule(lr){6-7}\cmidrule(lr){8-9}
    & RW $\uparrow$ & SR (\%) $\uparrow$
    & RW $\uparrow$ & SR (\%) $\uparrow$
    & RW $\uparrow$ & SR (\%) $\uparrow$
    & RW $\uparrow$ & SR (\%) $\uparrow$ \\
    \midrule
    IDM+MOBIL~\cite{Treiber2000IDM,Kesting2007MOBIL}
    & $-259.01 \pm 0.00$ & $49.42 \pm 0.00$
    & $-245.00 \pm 6.00$ & $10.77 \pm 0.25$
    & $-276.00 \pm 8.00$ & $20.67 \pm 0.45$
    & $122.41 \pm 10.44$ & $\mathbf{99.07 \pm 1.42}$ \\
    SAC~\cite{Haarnoja2018SAC}
    & $420.99 \pm 43.93$ & $27.20 \pm 9.17$
    & $454.99 \pm 36.14$ & $73.40 \pm 6.75$
    & $384.99 \pm 41.09$ & $79.40 \pm 8.74$
    & $194.69 \pm 12.28$ & $84.33 \pm 1.67$ \\
    PPO~\cite{Schulman2017PPO}
    & $136.42 \pm 10.35$ & $6.26 \pm 3.12$
    & $151.42 \pm 8.51$ & $18.66 \pm 2.30$
    & $120.42 \pm 9.68$ & $14.16 \pm 2.97$
    & $146.23 \pm 38.15$ & $85.00 \pm 5.40$ \\
    DreamerV3~\cite{Hafner2023DreamerV3}
    & $802.48 \pm 109.65$ & $85.75 \pm 4.68$
    & $836.48 \pm 85.05$ & $85.64 \pm 0.83$
    & $768.48 \pm 96.69$ & $81.64 \pm 1.07$
    & $207.30 \pm 8.20$ & $87.30 \pm 6.40$ \\
    R2-Dreamer~\cite{Morihira2026R2Dreamer}
    & $861.14 \pm 66.20$ & $90.38 \pm 5.37$
    & $895.14 \pm 51.35$ & $91.78 \pm 3.73$
    & $827.14 \pm 58.38$ & $88.18 \pm 4.82$
    & $233.24 \pm 12.28$ & $91.27 \pm 1.67$ \\
    \textbf{LIDAR-AD (Ours)}
    & $\mathbf{936.41 \pm 21.04}$ & $\mathbf{94.80 \pm 3.90}$
    & $\mathbf{967.41 \pm 17.31}$ & $\mathbf{94.46 \pm 2.26}$
    & $\mathbf{919.41 \pm 19.68}$ & $\mathbf{92.36 \pm 2.93}$
    & $\mathbf{259.74 \pm 9.21}$ & $97.60 \pm 3.20$ \\
    \bottomrule
  \end{tabular}
  }
  \par\vspace{1.0\baselineskip}
  \parbox{0.98\textwidth}{\footnotesize \emph{Note:} RW, reward; SR, success rate. All values are reported as mean $\pm$ standard deviation over $N_{\mathrm{run}}$ independent training runs.}
  % 中文翻译：MetaDrive 和 nuPlan 派生场景综合驾驶性能。
\end{table*}

\vspace{8pt}

\subsubsection{Learning Efficiency and Stability}

\begin{figure}[t]
  \centering
  \includegraphics[width=\linewidth]{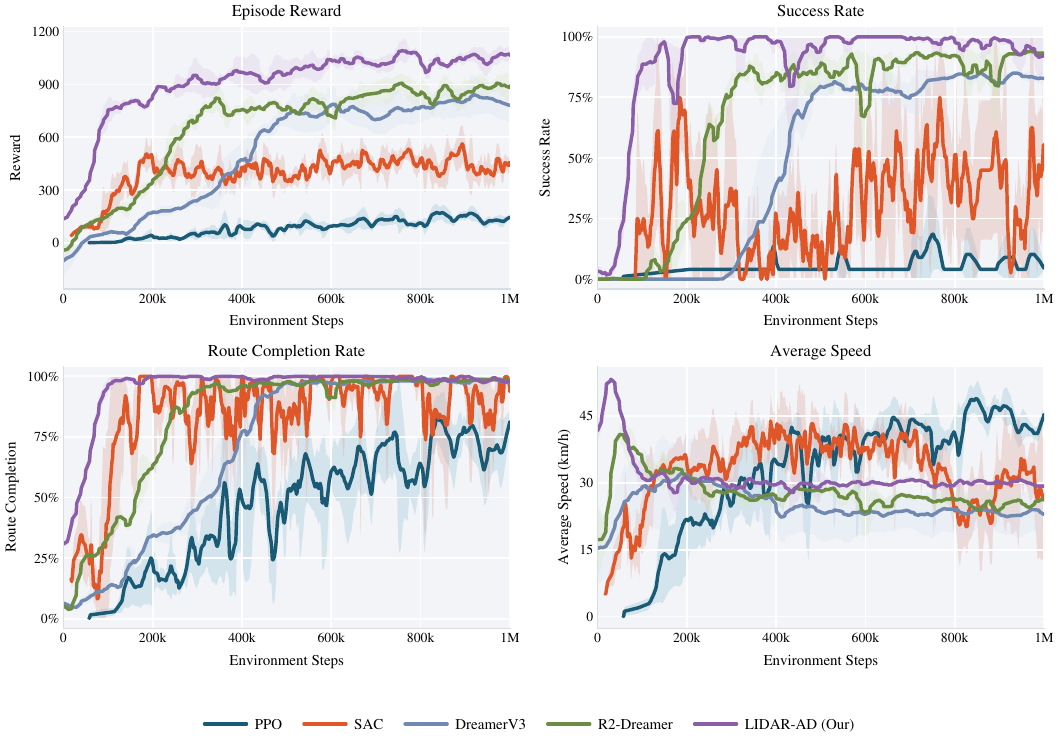}
  \caption{Training curves on the mixed MetaDrive scenario.
  LIDAR-AD reaches higher return and success levels with lower variance than the strongest world-model baselines, indicating improved learning stability under the same data budget.}
  \label{fig:training_curve}
  \vspace{-15pt}
\end{figure}

Fig.~\ref{fig:training_curve} compares training dynamics on the mixed MetaDrive scenario.
Under the same data budget, LIDAR-AD reaches higher return and success levels with lower variance than the strongest world-model baselines, indicating improved learning stability in addition to final closed-loop performance.
% 中文翻译：
% 图~\ref{fig:training_curve} 比较 mixed MetaDrive 场景上的训练过程。
% 在相同数据预算下，LIDAR-AD 相比最强世界模型基线达到更高 return 和 success，并表现出更低方差，说明其在提升最终闭环性能的同时也提升训练稳定性。

% 图片注释中文翻译：
% mixed MetaDrive 场景上的训练曲线。
% 相比最强世界模型基线，LIDAR-AD 达到更高的 return 和 success 水平，并表现出更低方差，说明其在相同数据预算下具有更稳定的学习过程。

\subsubsection{Module Contributions}

\begin{figure}[t]
  \centering
  \includegraphics[width=\linewidth]{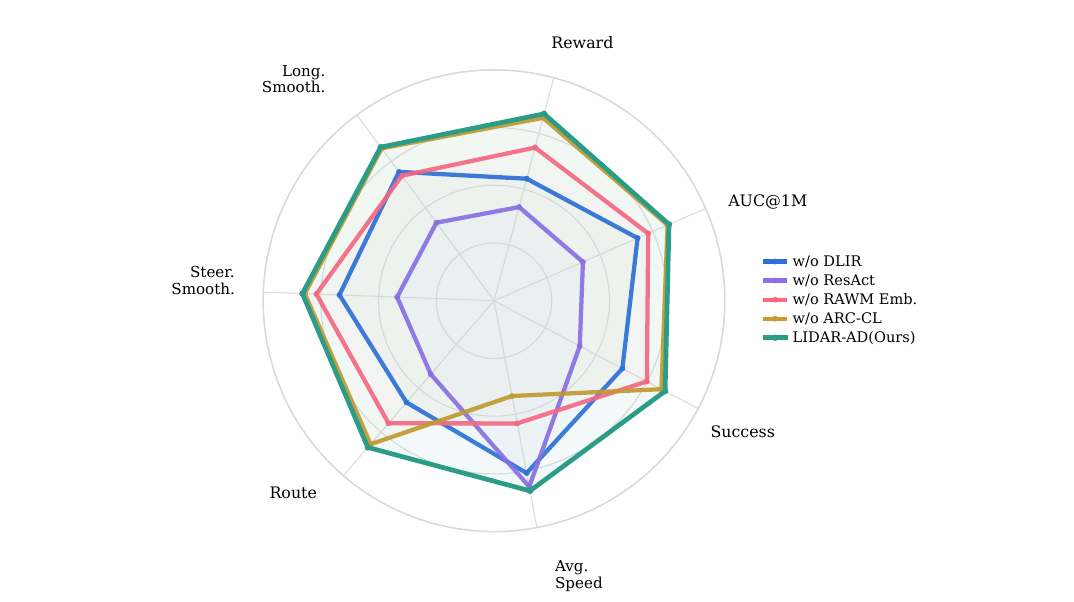}
  \caption{Ablation performance spectrum.
  The visualization summarizes the complementary effects of DLIR, ResAct, RAWM, and ARC-CL across driving performance, control smoothness, and sample efficiency.}
  \label{fig:ablation_spectrum}
  \vspace{-6pt}
\end{figure}

Table~\ref{tab:module_ablation} isolates the contribution of each module on the mixed MetaDrive scenario by removing one component from the full model at a time, and Fig.~\ref{fig:ablation_spectrum} provides a visual summary.
Removing DLIR reduces reward, success rate, route completion, control smoothness, and sample efficiency, showing the value of decoder-free latent interaction learning for state abstraction.
Removing ResAct produces the largest increase in steering and longitudinal-control deltas, while removing the RAWM embedding lowers task performance, indicating that smooth residual actions and residual-aware transitions provide complementary benefits.
Removing ARC-CL has smaller effects on control deltas but reduces reward, success rate, and route completion, showing that the multi-step contrastive objective mainly improves task completion and rollout consistency.
The full LIDAR-AD obtains the best reward, success rate, route completion, average speed, and steering smoothness.
% 中文翻译：
% 表~\ref{tab:module_ablation} 通过每次从完整模型中移除一个组件，分离各模块贡献；图~\ref{fig:ablation_spectrum} 给出可视化总结。
% 移除 DLIR 会降低 reward、success rate、route completion、控制平滑性和样本效率，说明无解码器潜在交互学习有助于状态抽象。
% 移除 ResAct 会显著增大 steering 和 longitudinal-control delta，而移除 RAWM embedding 会降低任务性能，说明平滑残差动作和残差感知转移具有互补作用。
% 移除 ARC-CL 对 action delta 的影响较小，但降低 reward、success rate 和 route completion，说明其主要改善任务完成和 rollout 一致性。
% 完整 LIDAR-AD 在 reward、success rate、route completion、average speed 和 steering smoothness 上取得最佳表现。

% 图片注释中文翻译：
% 消融性能谱。
% 该图总结了 DLIR、ResAct、RAWM 和 ARC-CL 在驾驶性能、控制平滑性和样本效率上的互补作用。

\begin{table*}[!t]
  \centering
  \caption{Module Ablation Study on the Mixed MetaDrive Scenario}
  \label{tab:module_ablation}
  \vspace{0.65\baselineskip}
  \setlength{\tabcolsep}{2.4pt}
  \footnotesize
  \resizebox{\textwidth}{!}{%
  \begin{tabular}{lccccccc}
    \toprule
    \textbf{Method} &
    \textbf{RW} $\uparrow$ &
    \textbf{SR (\%)} $\uparrow$ &
    \textbf{RC (\%)} $\uparrow$ &
    \textbf{AS} $\uparrow$ &
    \textbf{SD} $\downarrow$ &
    \textbf{LCD} $\downarrow$ &
    \textbf{nAUC@1M} $\uparrow$ \\
    \midrule
    LIDAR-AD w/o DLIR
    & $901.73 \pm 37.82$ & $91.46 \pm 3.28$ & $96.72 \pm 1.31$ & $28.86 \pm 0.74$
    & $0.047 \pm 0.007$ & $0.058 \pm 0.008$ & $0.812 \pm 0.013$ \\
    LIDAR-AD w/o ResAct
    & $895.62 \pm 42.38$ & $90.73 \pm 3.92$ & $96.41 \pm 1.42$ & $28.94 \pm 0.87$
    & $0.057 \pm 0.009$ & $0.071 \pm 0.010$ & $0.781 \pm 0.015$ \\
    LIDAR-AD w/o RAWM Emb.
    & $908.46 \pm 34.51$ & $91.88 \pm 3.41$ & $96.95 \pm 1.18$ & $28.57 \pm 0.69$
    & $0.043 \pm 0.006$ & $0.059 \pm 0.008$ & $0.818 \pm 0.012$ \\
    LIDAR-AD w/o ARC-CL
    & $914.87 \pm 28.53$ & $92.13 \pm 1.86$ & $97.18 \pm 0.97$ & $28.41 \pm 0.61$
    & $0.041 \pm 0.005$ & $0.052 \pm 0.006$ & $0.829 \pm 0.009$ \\
    \textbf{LIDAR-AD (Ours)}
    & $\mathbf{936.41 \pm 21.04}$ & $\mathbf{94.80 \pm 3.90}$ & $\mathbf{98.20 \pm 0.72}$ & $\mathbf{30.02 \pm 0.23}$
    & $\mathbf{0.035 \pm 0.004}$ & $0.051 \pm 0.007$ & $0.832 \pm 0.011$ \\
    \bottomrule
  \end{tabular}
  }
  \par\vspace{1.2\baselineskip}
  \parbox{0.98\textwidth}{\footnotesize \emph{Note:} Each row removes one component from the full LIDAR-AD model. ResAct denotes latent-tanh residual action parameterization, and RAWM Emb. denotes the residual-action-aware world-model embedding. RW, reward; SR, success rate; RC, route completion; AS, average speed; SD, steering delta; LCD, longitudinal-control delta; nAUC@1M, normalized AUC at 1M steps. All values are reported as mean $\pm$ standard deviation over $N_{\mathrm{run}}$ independent training runs.}
  % 中文翻译：每行从完整 LIDAR-AD 模型中移除一个组件；结果为 $N_{\mathrm{run}}$ 个独立训练 runs 的 mean ± std。
\end{table*}

\subsubsection{ARC-CL Horizon Sensitivity}

\begin{figure}[t]
  \centering
  \includegraphics[width=\linewidth]{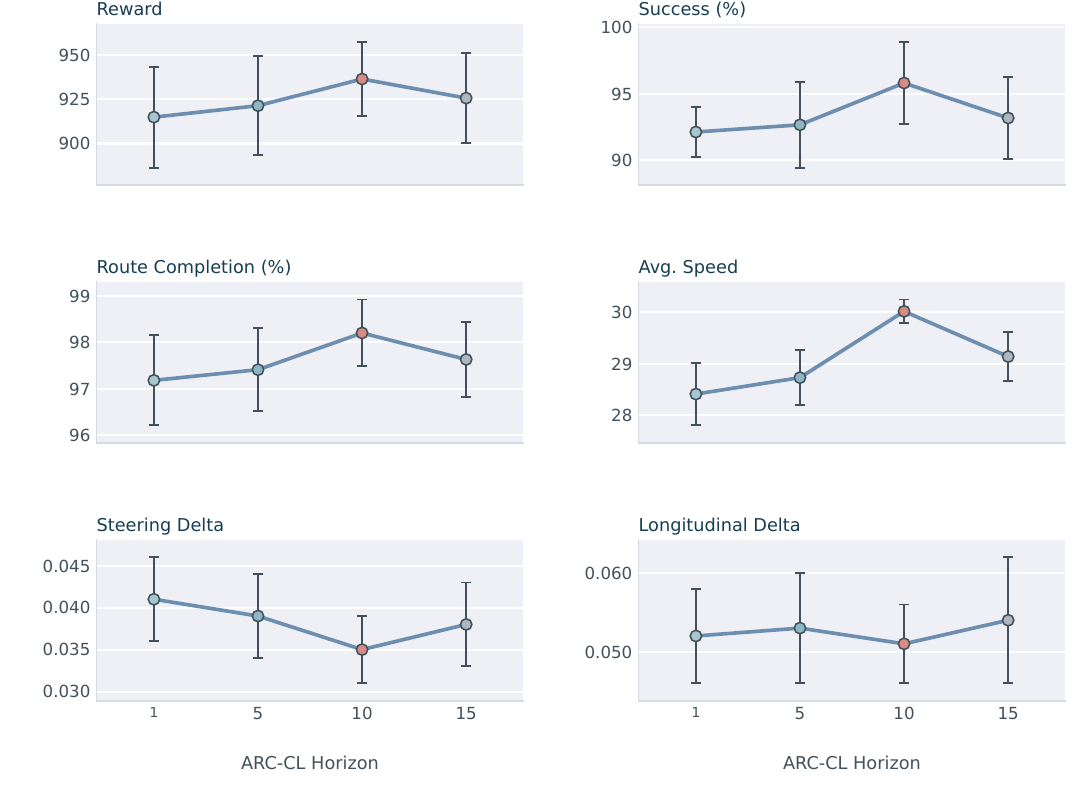}
  \caption{Effect of the ARC-CL rollout horizon.
  A ten-step horizon provides the best aggregate driving performance and the smoothest control among the evaluated settings.}
  \label{fig:arc_horizon}
  \vspace{-6pt}
\end{figure}

Fig.~\ref{fig:arc_horizon} examines the rollout horizon used by ARC-CL while keeping the remaining components fixed.
Among the evaluated horizons, $K=10$ achieves the best aggregate profile, with the highest reward, success rate, route completion, and average speed, as well as the lowest steering and longitudinal-control deltas.
A one-step objective provides limited temporal supervision, whereas $K=15$ slightly weakens driving performance and control smoothness.
We therefore use $K=10$ by default to balance multi-step control modeling and rollout error accumulation.
% 中文翻译：
% ARC-CL 时域长度的影响。
% 图~\ref{fig:arc_horizon} 在其余组件固定时考察 ARC-CL 的 rollout 时域。
% 在所评估的时域中，$K=10$ 取得最佳综合表现，包括最高 reward、success rate、route completion 和平均速度，
% 同时获得最低的转向和纵向控制变化量。
% 单步目标提供的时序监督有限，而 $K=15$ 会轻微削弱驾驶性能和控制平滑性。
% 因此，本文采用 $K=10$ 作为默认设置，以平衡多步控制建模与 rollout 误差累积。

% 图片注释中文翻译：
% ARC-CL rollout 时域的影响。
% 在所评估的设置中，十步时域取得了最佳综合驾驶性能和最平滑的控制。

\subsubsection{Traffic-Density Generalization from the Nominal Density}

\begin{figure*}[!b]
  \centering
  \includegraphics[width=\textwidth]{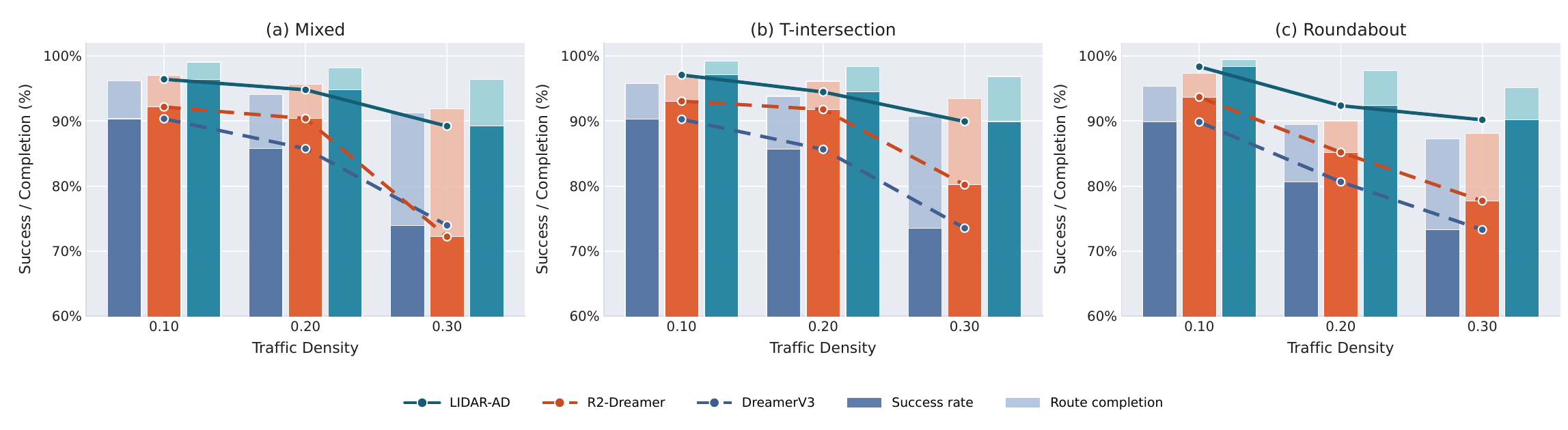}
  \caption{Traffic-density stress test.
  The grouped performance bars summarize mixed, T-intersection, and roundabout scenarios evaluated at traffic densities of 0.10, 0.20, and 0.30 after training at the nominal density of 0.20, where LIDAR-AD (Ours) maintains higher success rate and route completion than DreamerV3 and R2-Dreamer.}
  \label{fig:ood_density}
  \vspace{-4pt}
\end{figure*}

Table~\ref{tab:generalization} evaluates traffic-density generalization by training at the nominal density of 0.20 and testing mixed, T-intersection, and roundabout scenarios at densities of 0.10, 0.20, and 0.30.
All methods degrade as density increases, confirming that dense traffic is a harder closed-loop decision problem.
LIDAR-AD nevertheless achieves the highest success rate and route completion across all scenario-density pairs.
At density 0.30, it maintains success rates of 89.21\%, 89.92\%, and 88.18\% in mixed, T-intersection, and roundabout scenarios, whereas R2-Dreamer drops to 72.20\%, 80.18\%, and 80.72\%.
Together with Fig.~\ref{fig:ood_density}, these results indicate improved robustness under high-variation traffic flow rather than only fitting the nominal training density.
% 中文翻译：
% 表~\ref{tab:generalization} 在训练密度 0.20 下训练，并在 0.10、0.20、0.30 三个密度下测试 mixed、T-intersection 和 roundabout 场景，评估交通密度泛化。
% 随着密度增加，所有方法性能下降，说明密集交通是更困难的闭环决策问题。
% LIDAR-AD 仍在所有场景-密度组合上取得最高成功率和路线完成率。
% 在密度 0.30 下，LIDAR-AD 在三个场景中保持 89.21\%、89.92\% 和 88.18\% 的成功率，而 R2-Dreamer 分别下降到 72.20\%、80.18\% 和 80.72\%。
% 结合图~\ref{fig:ood_density}，这些结果说明该方法提升了高变化交通流下的鲁棒性，而不只是拟合标称训练密度。

% 图片注释中文翻译：
% 交通密度压力测试。
% 组图汇总在训练密度 0.20 后，mixed、T-intersection 和 roundabout 场景在 0.10、0.20、0.30 测试密度下的表现。
% LIDAR-AD (Ours) 在 success rate 和 route completion 上均高于 DreamerV3 和 R2-Dreamer。

\begin{table*}[!t]
  \centering
  \caption{Traffic-Density Generalization Across MetaDrive Scenarios}
  \label{tab:generalization}
  \setlength{\tabcolsep}{2.6pt}
  \scriptsize
  \resizebox{\textwidth}{!}{%
  \begin{tabular}{llcccccc}
    \toprule
    \multirow{2}{*}{\textbf{Scenario}} &
    \multirow{2}{*}{\textbf{Density}} &
    \multicolumn{2}{c}{\textbf{DreamerV3}~\cite{Hafner2023DreamerV3}} &
    \multicolumn{2}{c}{\textbf{R2-Dreamer}~\cite{Morihira2026R2Dreamer}} &
    \multicolumn{2}{c}{\textbf{LIDAR-AD (Ours)}} \\
    \cmidrule(lr){3-4}\cmidrule(lr){5-6}\cmidrule(lr){7-8}
    & & SR (\%) $\uparrow$ & RC (\%) $\uparrow$
      & SR (\%) $\uparrow$ & RC (\%) $\uparrow$
      & SR (\%) $\uparrow$ & RC (\%) $\uparrow$ \\
    \midrule
    \multirow{3}{*}{Mixed}
      & $0.10$ & $90.34 \pm 2.48$ & $96.18 \pm 1.35$ & $92.16 \pm 1.92$ & $97.02 \pm 1.11$ & $\mathbf{96.42 \pm 0.46}$ & $\mathbf{99.05 \pm 0.38}$ \\
      & $0.20$ & $85.75 \pm 4.68$ & $94.11 \pm 2.06$ & $90.38 \pm 5.37$ & $95.63 \pm 1.58$ & $\mathbf{94.80 \pm 3.90}$ & $\mathbf{98.20 \pm 0.72}$ \\
      & $0.30$ & $73.95 \pm 4.52$ & $91.26 \pm 2.84$ & $72.20 \pm 3.74$ & $91.88 \pm 2.41$ & $\mathbf{89.21 \pm 0.96}$ & $\mathbf{96.42 \pm 0.69}$ \\
    \midrule
    \multirow{3}{*}{T-Intersection}
      & $0.10$ & $90.26 \pm 3.06$ & $95.74 \pm 1.54$ & $93.05 \pm 2.15$ & $97.15 \pm 0.96$ & $\mathbf{97.08 \pm 0.64}$ & $\mathbf{99.21 \pm 0.31}$ \\
      & $0.20$ & $85.64 \pm 0.83$ & $93.82 \pm 1.03$ & $91.78 \pm 3.73$ & $96.11 \pm 1.42$ & $\mathbf{94.46 \pm 2.26}$ & $\mathbf{98.37 \pm 0.64}$ \\
      & $0.30$ & $73.52 \pm 4.88$ & $90.74 \pm 2.76$ & $80.18 \pm 3.96$ & $93.48 \pm 2.13$ & $\mathbf{89.92 \pm 1.05}$ & $\mathbf{96.85 \pm 0.73}$ \\
    \midrule
    \multirow{3}{*}{Roundabout}
      & $0.10$ & $89.84 \pm 4.15$ & $95.36 \pm 1.91$ & $93.68 \pm 3.28$ & $97.31 \pm 1.36$ & $\mathbf{98.35 \pm 0.93}$ & $\mathbf{99.42 \pm 0.27}$ \\
      & $0.20$ & $81.64 \pm 1.07$ & $92.48 \pm 1.46$ & $88.18 \pm 4.82$ & $95.04 \pm 1.72$ & $\mathbf{92.36 \pm 2.93}$ & $\mathbf{97.71 \pm 0.81}$ \\
      & $0.30$ & $73.28 \pm 6.72$ & $90.22 \pm 3.18$ & $80.72 \pm 5.67$ & $93.12 \pm 2.38$ & $\mathbf{88.18 \pm 1.94}$ & $\mathbf{95.16 \pm 0.92}$ \\
    \bottomrule
  \end{tabular}
  }
  \par\vspace{1.0\baselineskip}
  \parbox{0.98\textwidth}{\footnotesize \emph{Note:} SR, success rate; RC, route completion. All values are reported as mean $\pm$ standard deviation over $N_{\mathrm{run}}$ independent training runs.}
  % 中文翻译：三种方法在三个场景、三个密度等级下的成功率和路线完成率。
\end{table*}

\section{Discussion}
\label{sec:discussion}

% 写作指导：Discussion 不重复逐项结果，而是解释结果为何成立、对 TKDE 主题的意义以及边界条件。
\subsection{Mechanistic Interpretation}

The experimental results support the central design principle of LIDAR-AD: representation learning, residual control, and multi-step dynamics alignment should be optimized around the same RSSM latent state $s_t=(h_t,z_t)$.
Across mixed scene, roundabout, T-intersection, and nuPlan-derived log-reconstructed scenarios, LIDAR-AD improves the average success rate from 90.40\% for R2-Dreamer to 94.81\%, while achieving the highest reward in every scenario.
The component-removal ablation further shows that these gains do not come from a single isolated component.
Removing DLIR, ResAct, RAWM embedding, or ARC-CL consistently weakens at least one part of the performance profile, indicating that compact representation learning, residual control parameterization, action-aware dynamics, and multi-step rollout alignment provide complementary support for closed-loop driving.
% 中文翻译：实验结果支持 LIDAR-AD 的核心设计原则：表征学习、残差控制和多步动力学对齐应围绕同一个 RSSM 潜在状态 $s_t=(h_t,z_t)$ 优化。
% 在 mixed scene、roundabout、T-intersection 和 nuPlan-derived 日志重建场景中，LIDAR-AD 将平均成功率从 R2-Dreamer 的 90.40\% 提升到 94.81\%，并在每个场景中取得最高 reward。
% 组件移除消融进一步表明，这些收益并不来自某一个孤立组件。
% 移除 DLIR、ResAct、RAWM embedding 或 ARC-CL 都会持续削弱至少一个性能维度，说明紧凑表征学习、残差控制参数化、动作感知动力学和多步 rollout 对齐为闭环驾驶提供互补支撑。

\begin{figure}[!t]
  \centering
  \includegraphics[width=\linewidth]{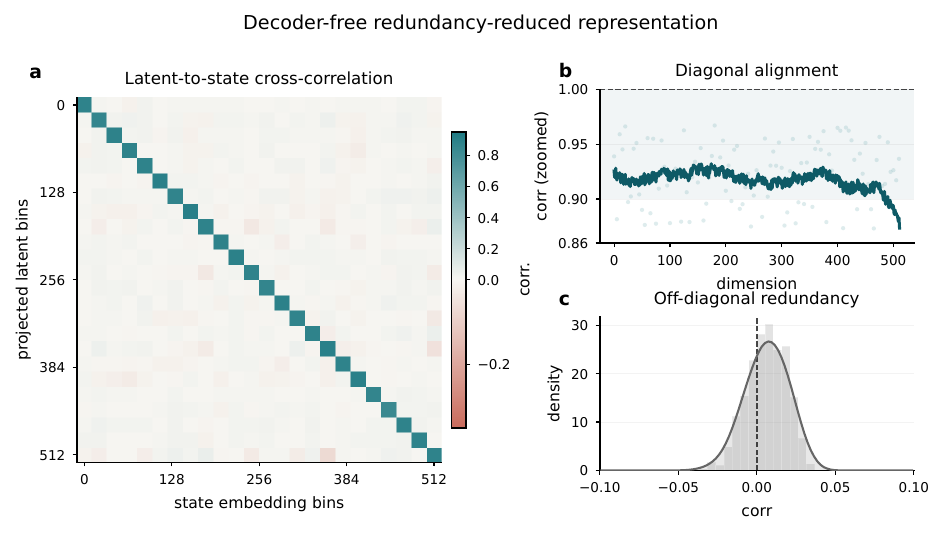}
  \caption{Latent representation structure.
  Cross-correlation matrices show that risk-guided decoder-free alignment reduces off-diagonal correlations, indicating a more compact latent representation for world-model learning.}
  \label{fig:barlow_representation}
  \vspace{-6pt}
\end{figure}

Fig.~\ref{fig:barlow_representation} provides a representation-level explanation for the performance gains.
Compared with reconstruction-based latent learning, the risk-guided decoder-free alignment produces a more diagonal cross-correlation pattern, indicating lower inter-dimensional redundancy.
This supports the role of DLIR as a capacity-allocation mechanism: the latent state is encouraged to preserve compact driving-relevant structure while avoiding supervision that scales mainly with raw modality size.
% 中文翻译：图~\ref{fig:barlow_representation} 从表征层面解释性能提升。
% 相比基于重建的潜在学习，风险引导的无解码器对齐产生更接近对角线的互相关模式，表明维度间冗余更低。
% 这支持将 DLIR 理解为一种容量分配机制：它鼓励潜在状态保留紧凑的驾驶相关结构，同时避免主要随原始模态尺寸扩大的监督信号。

% 图片注释中文翻译：
% 潜在表征结构。
% 互相关矩阵显示，风险引导的无解码器对齐降低了非对角相关性，说明世界模型学习到更紧凑的潜在表征。

\begin{figure}[!b]
  \centering
  \includegraphics[width=\linewidth]{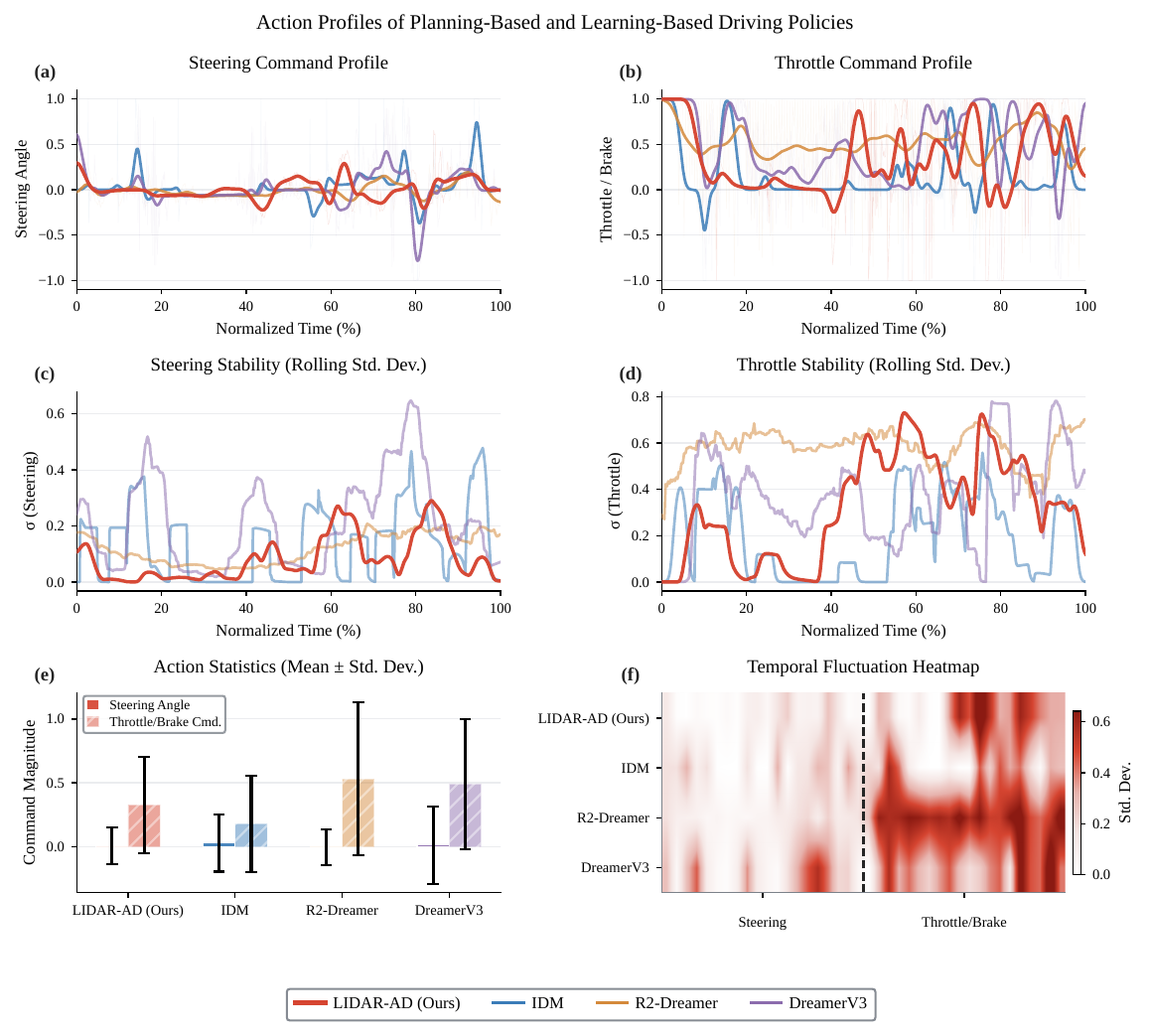}
  \caption{Temporal control profiles.
  Compared with absolute-action modeling, residual-action modeling produces smoother control evolution, consistent with the smooth-control bias introduced by RAWM.}
  \label{fig:action_smoothness}
  \vspace{-6pt}
\end{figure}

\begin{figure}[!t]
  \centering
  \includegraphics[width=\linewidth]{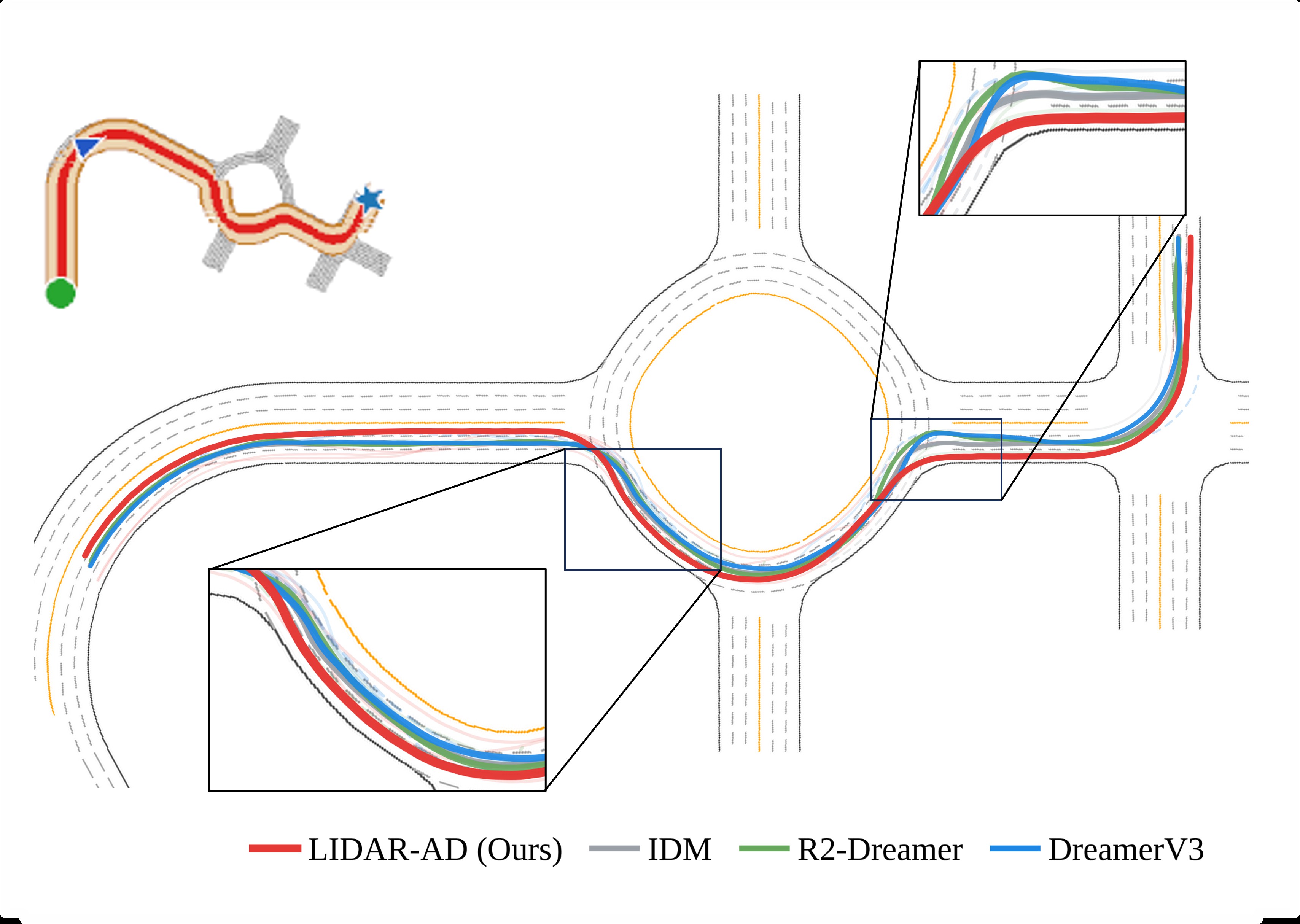}
  \caption{Representative closed-loop trajectories.
  LIDAR-AD follows the reference route more consistently through curved and roundabout segments, while the baselines show larger lateral deviations.}
  \label{fig:trajectory_comparison}
  \vspace{-4pt}
\end{figure}

Fig.~\ref{fig:action_smoothness} links the ablation evidence on steering and longitudinal deltas to actual control profiles.
The residual-action policy produces smoother temporal evolution than absolute-action modeling, which is consistent with the lower action deltas observed after introducing ResAct and RAWM.
Fig.~\ref{fig:trajectory_comparison} further suggests how this control regularity translates into closed-loop behavior: LIDAR-AD follows curved and roundabout routes more consistently, while the baselines show larger lateral deviations.
Together, these qualitative analyses explain why the quantitative gains appear simultaneously in success rate, route completion, reward, and traffic-density robustness.
% 中文翻译：图~\ref{fig:action_smoothness} 将转向和纵向控制 delta 的消融证据与实际控制曲线联系起来。
% 相比绝对动作建模，残差动作策略产生更平滑的时间演化，这与引入 ResAct 和 RAWM 后观察到的更低动作 delta 一致。
% 图~\ref{fig:trajectory_comparison} 进一步说明这种控制正则性如何转化为闭环行为：LIDAR-AD 在弯道和环岛路线中跟踪更一致，而基线方法表现出更大的横向偏差。
% 这些定性分析共同解释了为什么定量收益会同时体现在成功率、路线完成率、reward 和交通密度鲁棒性上。

% 图片注释中文翻译：
% 时序控制曲线。
% 相比绝对动作建模，残差动作建模产生更平滑的控制演化，与 RAWM 引入的平滑控制偏置一致。

\subsection{Implications for Structured and Temporal Knowledge Learning}

From a structured knowledge representation perspective, DLIR addresses the challenge of organizing heterogeneous driving observations into a compact latent state.
Ego kinematics, LiDAR-like range sensing, navigation context, and risk descriptors differ in scale, temporal behavior, and relevance to control.
By encoding these groups separately, modeling pairwise interactions, and reducing latent redundancy without observation reconstruction, DLIR provides a structured route for fusing heterogeneous data into a decision-oriented representation.
This allows the world model to organize heterogeneous driving signals into a compact control-relevant latent space without forcing the representation to reconstruct every observation detail.
% 中文翻译：从结构化知识表征角度看，DLIR 解决了将异构驾驶观测组织为紧凑潜在状态的问题。
% 自车运动学、类 LiDAR 距离感知、导航上下文和风险描述符在尺度、时间行为和控制相关性上各不相同。
% 通过分别编码这些组、建模成对交互，并在不重建观测的情况下减少潜在冗余，DLIR 为将异构数据融合为决策导向表征提供了结构化路径。
% 这使世界模型能够将异构驾驶信号组织到紧凑的控制相关潜在空间中，而不必强制表征重建每一个观测细节。

From a temporal knowledge learning perspective, RAWM and ARC-CL make action evolution part of the learned dynamics rather than treating each command as an independent absolute value.
RAWM supplies the transition model with both the executed command and its local residual change, while ARC-CL aligns the future state reached by recursively applying a sequence of residual-action embeddings.
This differs from single-step temporal contrastive learning such as CPC~\cite{Oord2018CPC}: the contrastive target is the latent state reached after a control sequence, not merely the next observation or next latent state.
For autonomous driving, this means the model is encouraged to distinguish action sequences that lead to different maneuver-level consequences, such as stable lane following, route deviation, or gap acceptance.
% 中文翻译：从时序知识学习角度看，RAWM 和 ARC-CL 将动作演化纳入已学习的动力学，而不是把每个命令视为独立的绝对值。
% RAWM 向转移模型同时提供已执行命令及其局部残差变化，而 ARC-CL 对齐通过递归应用一段 residual-action embedding 序列到达的未来状态。
% 这不同于 CPC 等单步时序对比学习：对比目标是经过控制序列后到达的潜在状态，而不仅是下一观测或下一潜在状态。
% 对自动驾驶而言，这意味着模型被鼓励区分会导致不同机动级后果的动作序列，例如稳定车道保持、路线偏离或间隙接受。

\begin{figure}[!t]
  \centering
  \includegraphics[width=\linewidth]{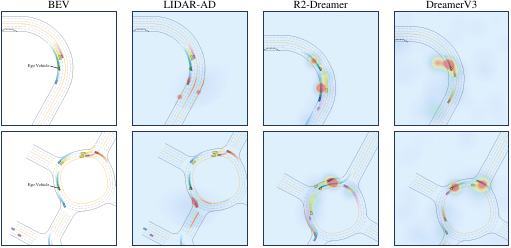}
  \caption{Policy saliency over bird's-eye-view driving layouts.
  The BEV column shows the scene layout, while warmer regions indicate higher policy saliency for each world-model policy.
  Compared with R2-Dreamer and DreamerV3, LIDAR-AD produces more localized responses around ego-relevant lane and interaction regions, suggesting stronger alignment between the learned latent dynamics and control-relevant cues.}
  \label{fig:policy_saliency}
  \vspace{-6pt}
\end{figure}

Fig.~\ref{fig:policy_saliency} provides a complementary policy-level diagnostic of the learned world model.
Across both driving layouts, LIDAR-AD concentrates saliency around the ego-relevant lane corridor, nearby agents, and conflict regions, whereas R2-Dreamer and DreamerV3 show more diffuse or spatially shifted responses.
This pattern is consistent with the intended role of DLIR and RAWM: the latent state is guided toward interaction and residual-control cues that directly affect the imagined transition and policy update.
It helps explain why LIDAR-AD improves over other world-model baselines in closed-loop reward, success rate, and density generalization.
% 中文翻译：图~\ref{fig:policy_saliency} 从策略层面补充诊断学习到的世界模型。
% 在两类驾驶布局中，LIDAR-AD 的 saliency 更集中于自车相关车道走廊、附近交通参与者和冲突区域，而 R2-Dreamer 和 DreamerV3 的响应更分散或出现空间偏移。
% 这一现象与 DLIR 和 RAWM 的设计作用一致：潜在状态被引导关注会直接影响想象转移和策略更新的交互线索与残差控制线索。
% 这有助于解释 LIDAR-AD 为什么在闭环 reward、success rate 和密度泛化上优于其他世界模型基线。

% 图片注释中文翻译：
% 鸟瞰驾驶布局上的策略 saliency。
% BEV 列显示场景布局，颜色越暖表示对应世界模型策略的 saliency 越高。
% 相比 R2-Dreamer 和 DreamerV3，LIDAR-AD 在自车相关车道和交互区域附近产生更局部化的响应，说明其学习到的潜在动力学与控制相关线索更一致。

% 图片注释中文翻译：
% 代表性闭环轨迹。
% LIDAR-AD 在弯道和环岛路段中更稳定地跟随参考路线，而基线方法表现出更大的横向偏差。

\subsection{Limitations and Future Directions}

The present evaluation is still bounded by simulator-based closed-loop testing and nuPlan-derived scenarios, which cannot fully reproduce real sensor noise, calibration error, weather variation, localization drift, or vehicle dynamics. LIDAR-AD also uses a fixed ARC-CL horizon of $K=10$, pairwise modality interactions, and compact LiDAR-like observations, leaving adaptive horizon selection, higher-order interaction modeling, and extension to denser sensing streams as natural next steps. Finally, although the method improves reward, success rate, route completion, and control smoothness, it does not yet impose explicit safety constraints inside latent imagination; future work should therefore prioritize real-world closed-loop validation and safety-aware latent planning.
% 中文翻译：当前评估仍受限于基于仿真的闭环测试和 nuPlan-derived 场景，无法完全复现真实传感器噪声、标定误差、天气变化、定位漂移或车辆动力学。
% LIDAR-AD 还使用固定的 ARC-CL 时域 $K=10$、成对模态交互和紧凑的类 LiDAR 观测，因此自适应时域选择、高阶交互建模以及扩展到更稠密感知流是自然的后续方向。
% 最后，尽管该方法提升了 reward、成功率、路线完成率和控制平滑性，但尚未在潜在想象中施加显式安全约束；因此，未来工作应优先关注真实世界闭环验证和安全感知潜在规划。

% ===========================================================================
% VIII. CONCLUSION
% 写作指导：总结方法贡献，呼应未来工作。

\section{Conclusion}
\label{sec:conclusion}

% 写作指导：总结核心方法及三大模块。
This paper has presented \textbf{LIDAR-AD}, a decoder-free latent-interaction dreamer world model with residual-action rollout alignment for autonomous driving in simulated environments.
LIDAR-AD addresses three fundamental limitations of existing world model approaches to autonomous driving:
% 中文翻译：本文提出了 \textbf{LIDAR-AD}，一种面向仿真自动驾驶、带有残差动作 rollout 对齐的无解码器潜在交互 Dreamer 世界模型。
% LIDAR-AD 解决现有自动驾驶世界模型方法中的三个根本局限。
% 写作指导：三局限性对应三模块。
\begin{enumerate}[leftmargin=*, itemsep=0.5ex]
  \item The inefficiency of full observation reconstruction for heterogeneous driving modalities, addressed by the \textbf{Decoder-Free Latent Interaction Representation (DLIR)} module, which learns compact latent states through risk-guided grouped encoding, structured pairwise interaction modeling, and redundancy reduction---without requiring an observation decoder.
  % 中文翻译：针对异构驾驶模态中完整观测重建效率较低的问题，\textbf{无解码器潜在交互表征（DLIR）}通过风险引导分组编码、结构化成对交互建模和冗余消减来学习紧凑潜在状态，并且不需要观测解码器。
  \item The lack of temporal continuity in absolute action prediction, addressed by the \textbf{Residual-Action World Model (RAWM)}, which uses latent-tanh residual actions to preserve interior action reachability, keep zero-residual commands invariant, and enrich the dynamics transition with joint action context.
  % 中文翻译：针对绝对动作预测缺少时间连续性的问题，\textbf{残差动作世界模型（RAWM）}使用 latent-tanh 残差动作，在保持动作内部可达性和零残差命令不变性的同时，用联合动作上下文增强动力学转移。
  \item The inability to capture cumulative consequences of continuous action sequences, addressed by \textbf{Residual-Action Chain Contrastive Learning (ARC-CL)}, which aligns multi-step prior rollouts recursively driven by per-step residual-action inputs with true future posterior states.
  % 中文翻译：针对难以捕获连续动作序列累积后果的问题，\textbf{残差动作链对比学习（ARC-CL）}将由逐步 residual-action 输入递归驱动的多步先验 rollout 与真实未来后验状态对齐。
\end{enumerate}
These three modules operate on a shared RSSM latent state $s_t$, forming a tightly integrated system where improvements in representation, control, and dynamics prediction mutually reinforce each other.
The residual-action statistics in Fig.~\ref{fig:residual_action_horizon} and the latent-tanh analysis further characterize why RAWM can express smooth control as compact residual updates without restricting the interior executable action range.
% 中文翻译：这三个模块共同作用于共享 RSSM 潜在状态 $s_t$，形成一个表征、控制和动力学预测相互增强的紧密集成系统。
% 图~\ref{fig:residual_action_horizon} 中的残差动作统计和 latent-tanh 分析进一步说明，RAWM 为什么能够在不限制内部可执行动作范围的情况下，将平滑控制表示为紧凑的残差更新。
% 写作指导：呼应协同增益

Experiments on mixed scene, roundabout, T-intersection, and nuPlan-derived log-reconstructed scenarios show that LIDAR-AD achieves the highest reward in all scenarios and the best success rate among learning-based methods.
Its average success rate reaches 94.81\%, compared with 90.40\% for R2-Dreamer and 85.08\% for DreamerV3.
These results indicate that the proposed risk-guided representation, residual-action modeling, and multi-step residual-action contrastive objective provide complementary benefits under a unified world-model framework.
% 中文翻译：在 mixed scene、roundabout、T-intersection 和 nuPlan-derived 日志重建场景上的实验表明，LIDAR-AD 在所有场景中取得最高 reward，并在学习式方法中取得最佳成功率。
% 其平均成功率达到 94.81\%，相比之下 R2-Dreamer 为 90.40\%，DreamerV3 为 85.08\%。
% 这些结果表明，所提出的风险引导表征、残差动作建模和多步残差动作对比目标在统一世界模型框架下具有互补收益。

% 写作指导：未来工作方向，避免与 Discussion 中的局限性列表重复。
Future work will prioritize real-world closed-loop validation and adaptive, safety-aware latent planning.
% 中文翻译：未来工作将优先关注真实世界闭环验证以及自适应、安全感知的潜在规划。

% ===========================================================================
% 参考文献
% ===========================================================================
% 写作指导：复用已验证引用，不生成虚假文献。

% ===========================================================================
% 参考文献
% ===========================================================================
\bibliographystyle{IEEEtran}
\bibliography{references}

@article{Paden2016Survey,
	author    = {Brian Paden and Michal {\v{C}}{\'{a}}p and Sze Zheng Yong and Dmitry Yershov and Emilio Frazzoli},
	title     = {A Survey of Motion Planning and Control Techniques for Self-Driving Urban Vehicles},
	journal   = {IEEE Trans. Intell. Veh.},
	volume    = {1},
	number    = {1},
	pages     = {33--55},
	year      = {2016},
	publisher = {IEEE}
}

@article{Schulman2017PPO,
	author    = {John Schulman and Filip Wolski and Prafulla Dhariwal and Alec Radford and Oleg Klimov},
	title     = {Proximal Policy Optimization Algorithms},
	journal   = {arXiv preprint arXiv:1707.06347},
	year      = {2017}
}

@inproceedings{Haarnoja2018SAC,
	author    = {Tuomas Haarnoja and Aurick Zhou and Pieter Abbeel and Sergey Levine},
	title     = {Soft Actor-Critic: Off-Policy Maximum Entropy Deep Reinforcement Learning with a Stochastic Actor},
	booktitle = {Proc. Int. Conf. Mach. Learn. (ICML)},
	year      = {2018},
	pages     = {1861--1870}
}

@inproceedings{Hafner2019PlaNet,
	author    = {Danijar Hafner and Timothy Lillicrap and Ian Fischer and Ruben Villegas and David Ha and Honglak Lee and James Davidson},
	title     = {Learning Latent Dynamics for Planning from Pixels},
	booktitle = {Proc. Int. Conf. Mach. Learn. (ICML)},
	year      = {2019},
	pages     = {2555--2565}
}

@inproceedings{Hafner2020Dreamer,
	author    = {Danijar Hafner and Timothy Lillicrap and Jimmy Ba and Mohammad Norouzi},
	title     = {Dream to Control: Learning Behaviors by Latent Imagination},
	booktitle = {Proc. Int. Conf. Learn. Representations (ICLR)},
	year      = {2020}
}

@article{Hafner2023DreamerV3,
	author    = {Danijar Hafner and Jurgis Pasukonis and Jimmy Ba and Timothy Lillicrap},
	title     = {Mastering Diverse Domains through World Models},
	journal   = {arXiv preprint arXiv:2301.04104},
	year      = {2023}
}

@article{Gao2024SEM2,
	author    = {Zizhang Gao and Yao Mu and Cheng Chen and Jingyuan Duan and Ping Luo and Yang Lu and Shengbo Eben Li},
	title     = {Enhance Sample Efficiency and Robustness of End-to-End Urban Autonomous Driving via Semantic Masked World Model},
	journal   = {IEEE Trans. Intell. Transp. Syst.},
	volume    = {25},
	number    = {6},
	pages     = {5678--5692},
	year      = {2024},
	publisher = {IEEE}
}

@inproceedings{Huang2024SafeDreamer,
	author    = {Weidong Huang and Jiaming Ji and Chunhe Xia and Borui Zhang and Yaodong Yang},
	title     = {SafeDreamer: Safe Reinforcement Learning with World Models},
	booktitle = {Proc. Int. Conf. Learn. Representations (ICLR)},
	year      = {2024}
}

@inproceedings{Janner2019MBPO,
	author    = {Michael Janner and Justin Fu and Marvin Zhang and Sergey Levine},
	title     = {When to Trust Your Model: Model-Based Policy Optimization},
	booktitle = {Adv. Neural Inf. Process. Syst. (NeurIPS)},
	year      = {2019},
	pages     = {12519--12530}
}

@inproceedings{Liu2025ITSC,
	author    = {Yongzhi Liu and Sunan Zhang and Changfeng Shen and Xiangwei Zhang and Weichao Zhuang},
	title     = {Risk-Aware Dual-Policy Coordination with World Model for Safe and Adaptive Highway Autonomous Driving},
	booktitle = {Proc. IEEE Intell. Transp. Syst. Conf. (ITSC)},
	year      = {2025},
	pages     = {776--782},
	address   = {Gold Coast, Australia},
	month     = {Nov.},
	doi       = {10.1109/ITSC60802.2025.11423709},
	note      = {doi: 10.1109/ITSC60802.2025.11423709}
}

@article{Treiber2000IDM,
	author    = {Martin Treiber and Ansgar Hennecke and Dirk Helbing},
	title     = {Congested Traffic States in Empirical Observations and Microscopic Simulations},
	journal   = {Phys. Rev. E},
	volume    = {62},
	number    = {2},
	pages     = {1805--1824},
	year      = {2000},
	publisher = {American Physical Society}
}

@article{Kesting2007MOBIL,
	author    = {Arne Kesting and Martin Treiber and Dirk Helbing},
	title     = {General Lane-Changing Model {MOBIL} for Car-Following Models},
	journal   = {Transp. Res. Rec.},
	volume    = {1999},
	number    = {1},
	pages     = {86--94},
	year      = {2007},
	publisher = {SAGE Publications}
}

@IEEEtranBSTCTL{IEEEtranBSTCTL:no_dash,
  CTLdash_repeated_names = "no"
}

@article{Kiran2022Deep,
  author  = {B. R. Kiran and Ibrahim Sobh and Victor Talpaert and Patrick Mannion and Ahmad A. Al Sallab and Senthil Yogamani and Patrick Perez},
  title   = {Deep Reinforcement Learning for Autonomous Driving: A Survey},
  journal = {IEEE Trans. Intell. Transp. Syst.},
  volume  = {23},
  number  = {6},
  pages   = {4909--4926},
  year    = {2022}
}

@inproceedings{Morihira2026R2Dreamer,
  author    = {Noboru Morihira and Adnan Nahar and Homanga Bharadwaj and Yusuke Kato and Akio Hayashi and Tatsuya Harada},
  title     = {{R2-Dreamer}: Redundancy-Reduced World Models Without Decoders or Augmentation},
  booktitle = {Proc. Int. Conf. Learn. Representations (ICLR)},
  year      = {2026},
  note      = {arXiv:2603.18202}
}

@inproceedings{Liu2025ResAct,
  author    = {Ziyu Liu and Peng Peng and Yuandong Tian},
  title     = {Visual Reinforcement Learning with Residual Action},
  booktitle = {Proc. AAAI Conf. Artif. Intell.},
  volume    = {39},
  number    = {18},
  pages     = {19050--19058},
  year      = {2025}
}

@article{Zhang2026ResWM,
  author  = {Jiawei Zhang and Gayan Adineera and Jie Tan and Jung Kim},
  title   = {{ResWM}: Residual-Action World Model for Visual {RL}},
  journal = {arXiv preprint arXiv:2603.11110},
  year    = {2026}
}

@inproceedings{Hansen2022TDMPC,
  author    = {Nicklas A. Hansen and Hao Su and Xiaolong Wang},
  title     = {Temporal Difference Learning for Model Predictive Control},
  booktitle = {Proc. Int. Conf. Mach. Learn. (ICML)},
  series    = {Proc. Mach. Learn. Res.},
  volume    = {162},
  pages     = {8387--8406},
  year      = {2022}
}

@inproceedings{Hansen2024TDMPC2,
  author    = {Nicklas A. Hansen and Hao Su and Xiaolong Wang},
  title     = {{TD-MPC2}: Scalable, Robust World Models for Continuous Control},
  booktitle = {Proc. Int. Conf. Learn. Representations (ICLR)},
  year      = {2024}
}

@inproceedings{Hu2022MILE,
  author    = {Anthony Hu and Gianluca Corrado and Nicholas Griffiths and Zac Murez and Catalin Gurau and Hudson Yeo and Alex Kendall and Roberto Cipolla and Jamie Shotton},
  title     = {Model-Based Imitation Learning for Urban Driving},
  booktitle = {Adv. Neural Inf. Process. Syst. (NeurIPS)},
  year      = {2022}
}

@inproceedings{Li2024Think2Drive,
  author    = {Quanyi Li and Xiaosong Jia and Shaobo Wang and Junchi Yan},
  title     = {{Think2Drive}: Efficient Reinforcement Learning by Thinking in Latent World Model for Quasi-Realistic Autonomous Driving},
  booktitle = {Proc. Eur. Conf. Comput. Vis. (ECCV)},
  year      = {2024}
}

@inproceedings{Wang2023DriveDreamer,
  author    = {Xiaofeng Wang and Zheng Zhu and Guan Huang and Xin Chen and Jie Zhu and Jiwen Lu},
  title     = {{DriveDreamer}: Towards Real-World-Driven World Models for Autonomous Driving},
  booktitle = {Proc. Eur. Conf. Comput. Vis. (ECCV)},
  year      = {2024}
}

@inproceedings{Zheng2024GenAD,
  author    = {Wenzhao Zheng and Ruiqi Song and Xianda Guo and Chenming Zhang and Long Chen},
  title     = {{GenAD}: Generative End-to-End Autonomous Driving},
  booktitle = {Proc. Eur. Conf. Comput. Vis. (ECCV)},
  year      = {2024}
}

@article{Oord2018CPC,
  author  = {Aaron van den Oord and Yazhe Li and Oriol Vinyals},
  title   = {Representation Learning with Contrastive Predictive Coding},
  journal = {arXiv preprint arXiv:1807.03748},
  year    = {2018}
}

@inproceedings{Grill2020BYOL,
  author    = {Jean-Bastien Grill and Florian Strub and Florent Altche and Corentin Tallec and Pierre H. Richemond and Elena Buchatskaya and Carl Doersch and Bernardo Avila Pires and Zhaohan Daniel Guo and Mohammad Gheshlaghi Azar and Bilal Piot and Koray Kavukcuoglu and Remi Munos and Michal Valko},
  title     = {Bootstrap Your Own Latent: A New Approach to Self-Supervised Learning},
  booktitle = {Adv. Neural Inf. Process. Syst. (NeurIPS)},
  volume    = {33},
  pages     = {21271--21284},
  year      = {2020}
}

@inproceedings{Zbontar2021Barlow,
  author    = {Jure Zbontar and Li Jing and Ishan Misra and Yann LeCun and Stephane Deny},
  title     = {Barlow Twins: Self-Supervised Learning via Redundancy Reduction},
  booktitle = {Proc. Int. Conf. Mach. Learn. (ICML)},
  series    = {Proc. Mach. Learn. Res.},
  volume    = {139},
  pages     = {12310--12320},
  year      = {2021}
}

@inproceedings{Assran2023IJEPA,
  author    = {Mahmoud Assran and Quentin Duval and Ishan Misra and Piotr Bojanowski and Pascal Vincent and Michael Rabbat and Yann LeCun and Nicolas Ballas},
  title     = {Self-Supervised Learning from Images with a Joint-Embedding Predictive Architecture},
  booktitle = {Proc. IEEE/CVF Conf. Comput. Vis. Pattern Recognit. (CVPR)},
  pages     = {15619--15629},
  year      = {2023}
}

@article{Bardes2024VJEPA,
  author  = {Adrien Bardes and Quentin Garrido and Jean Ponce and Xinlei Chen and Michael Rabbat and Yann LeCun and Mahmoud Assran and Nicolas Ballas},
  title   = {Revisiting Feature Prediction for Learning Visual Representations from Video},
  journal = {arXiv preprint arXiv:2404.08471},
  year    = {2024},
  note    = {v-JEPA}
}

@inproceedings{Laskin2020CURL,
  author    = {Michael Laskin and Aravind Srinivas and Pieter Abbeel},
  title     = {{CURL}: Contrastive Unsupervised Representations for Reinforcement Learning},
  booktitle = {Proc. Int. Conf. Mach. Learn. (ICML)},
  series    = {Proc. Mach. Learn. Res.},
  volume    = {119},
  pages     = {5639--5650},
  year      = {2020}
}

@inproceedings{Schwarzer2021SPR,
  author    = {Max Schwarzer and Ankesh Anand and Rishabh Goel and R. Devon Hjelm and Aaron Courville and Philip Bachman},
  title     = {Data-Efficient Reinforcement Learning with Self-Predictive Representations},
  booktitle = {Proc. Int. Conf. Learn. Representations (ICLR)},
  year      = {2021}
}

@inproceedings{Kostrikov2021DrQ,
  author    = {Denis Yarats and Ilya Kostrikov and Rob Fergus},
  title     = {Image Augmentation Is All You Need: Regularizing Deep Reinforcement Learning from Pixels},
  booktitle = {Proc. Int. Conf. Learn. Representations (ICLR)},
  year      = {2021}
}

@inproceedings{Yarats2022DrQv2,
  author    = {Denis Yarats and Rob Fergus and Alessandro Lazaric and Lerrel Pinto},
  title     = {Mastering Visual Continuous Control: Improved Data-Augmented Reinforcement Learning},
  booktitle = {Proc. Int. Conf. Learn. Representations (ICLR)},
  year      = {2022}
}

@inproceedings{Johannink2019ResidualRL,
  author    = {Tobias Johannink and Shikhar Bahl and Aravind Nair and Jianlan Luo and Avi Kumar and Matthias Loskyll and Juan Aparicio Ojea and Eugen Solowjow and Sergey Levine},
  title     = {Residual Reinforcement Learning for Robot Control},
  booktitle = {Proc. IEEE Int. Conf. Robot. Autom. (ICRA)},
  pages     = {6023--6029},
  year      = {2019}
}

@inproceedings{Mysore2021CAPS,
  author    = {Sanket Mysore and Bassel Mabsout and Renato Mancuso and Kate Saenko},
  title     = {Regularizing Action Policies for Smooth Control with Reinforcement Learning},
  booktitle = {Proc. IEEE Int. Conf. Robot. Autom. (ICRA)},
  year      = {2021}
}

@inproceedings{Chi2023DiffusionPolicy,
  author    = {Cheng Chi and Siyuan Feng and Yilun Du and Zhenjia Xu and Eric Cousineau and Benjamin Burchfiel and Shuran Song},
  title     = {Diffusion Policy: Visuomotor Policy Learning via Action Diffusion},
  booktitle = {Robotics: Science and Systems (RSS)},
  year      = {2023}
}

@inproceedings{Zhang2025CoA,
  author    = {Wenzhao Zhang and Tao Hu and Haozhe Zhang and Yanjie Qiao and Yuzhe Qin and Yifei Li and Jiajun Liu and Tao Kong and Li Liu and Xueqian Ma},
  title     = {Chain-of-Action: Trajectory Autoregressive Modeling for Robotic Manipulation},
  booktitle = {Adv. Neural Inf. Process. Syst. (NeurIPS)},
  year      = {2025}
}

@article{Li2022MetaDrive,
  author  = {Quanyi Li and Zhenghao Peng and Lan Feng and Qihang Zhang and Zhizheng Xue and Bolei Zhou},
  title   = {{MetaDrive}: Composing Diverse Driving Scenarios for Generalizable Reinforcement Learning},
  journal = {IEEE Trans. Pattern Anal. Mach. Intell.},
  volume  = {45},
  number  = {3},
  pages   = {3461--3475},
  year    = {2023},
  note    = {arXiv:2109.12674}
}

@article{Caesar2021NuPlan,
  author  = {Holger Caesar and Juraj Kabzan and Kok Seang Tan and Whye Kit Fong and Eric M. Wolff and Alex Lang and Luke Fletcher and Oscar Beijbom and Sammy Omari},
  title   = {{nuPlan}: A Closed-Loop {ML}-Based Planning Benchmark for Autonomous Vehicles},
  journal = {arXiv preprint arXiv:2106.11810},
  year    = {2021}
}

@inproceedings{Li2023ScenarioNet,
  author    = {Quanyi Li and Zhenghao Peng and Lan Feng and Zhizheng Liu and Chenda Duan and Wenjie Mo and Bolei Zhou},
  title     = {{ScenarioNet}: Open-Source Platform for Large-Scale Traffic Scenario Simulation and Modeling},
  booktitle = {Adv. Neural Inf. Process. Syst. (NeurIPS) Datasets and Benchmarks Track},
  year      = {2023}
}

\end{document}